\documentclass[10pt,twocolumn,letterpaper]{article}

\usepackage[pagenumbers]{cvpr}      
\usepackage{graphicx}
\usepackage{float}
\usepackage{booktabs}
\usepackage{amsmath}
\usepackage[misc]{ifsym}
\usepackage[linesnumbered,ruled]{algorithm2e}
 \usepackage{multirow}

\usepackage[pagebackref,breaklinks,colorlinks]{hyperref}
\usepackage[accsupp]{axessibility}

\def\cvprPaperID{1} 
\def\confName{CVPR}
\def\confYear{2022}

\begin{document}

\title{Delving into High-Quality Synthetic Face Occlusion Segmentation Datasets}

\author{Kenny T. R. Voo \hspace{12pt} Liming Jiang \hspace{12pt} Chen Change Loy\\[2pt]
S-Lab, Nanyang Technological University\\
{\tt\small \{kvoo001,liming002,ccloy\}@ntu.edu.sg} 
}

\maketitle

\begin{abstract}

This paper performs comprehensive analysis on datasets for occlusion-aware face segmentation, a task that is crucial for many downstream applications.
The collection and annotation of such datasets are time-consuming and labor-intensive. Although some efforts have been made in synthetic data generation, the naturalistic aspect of data remains less explored. In our study, we propose two occlusion generation techniques, Naturalistic Occlusion Generation (NatOcc), for producing high-quality naturalistic synthetic occluded faces; and Random Occlusion Generation (RandOcc), a more general synthetic occluded data generation method (Figure~\ref{fig:teaser}). We empirically show the effectiveness and robustness of both methods, even for unseen occlusions.
To facilitate model evaluation, we present two high-resolution real-world occluded face datasets with fine-grained annotations, RealOcc and RealOcc-Wild, featuring both careful alignment preprocessing and an in-the-wild setting for robustness test. We further conduct a comprehensive analysis on a newly introduced segmentation benchmark, offering insights for future exploration. Our code and dataset are available at \href{https://github.com/kennyvoo/face_occlusion_generation}{https://github.com/kennyvoo/face-occlusion-generation}.

\end{abstract}

\section{Introduction}
\label{sec:intro}

This paper explores the problem of occlusion-aware face segmentation, which involves extracting face pixels from an occluded face where occlusions are treated as the background class. Face or occlusion semantic segmentation is extensively used in face-related tasks, such as face recognition~\cite{occlusionpairwise}, face-swapping~\cite{nirkin2017face,perov2020deepfacelab}, and facial reconstruction~\cite{reconstruction, yin2021segmentationreconstructionguided}.
Occlusions on faces remained less explored in the past. In reality,
human faces are likely to be partially occluded by hands or wearable objects such as sunglasses or face masks.
These occlusions would often degrade the performance of the face-related tasks.
Therefore, occlusion-aware face segmentation is worth studying for many practical applications~\cite{nirkin2017face,dataset_jiang2020deeperforensics1,yin2021segmentationreconstructionguided}.

\begin{figure}[t]
  \centering
  \includegraphics[width=1\linewidth]{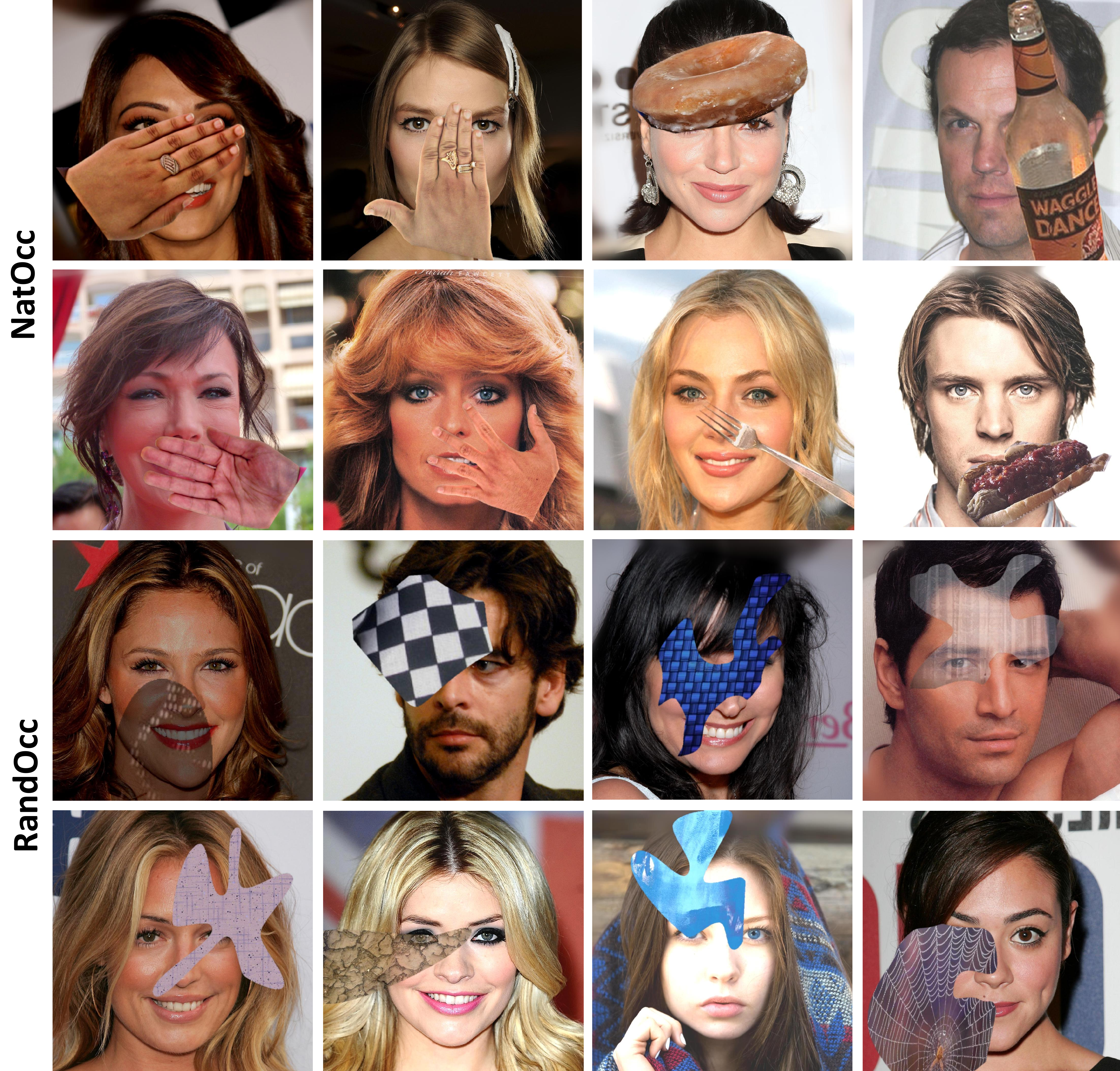}
  \vspace{-6mm}
  \caption{Examples of data generated by our NatOcc and RandOcc methods from CelebAMask-HQ~\cite{dataset_CelebAMask-HQ}. The first two rows are naturalistic occluded faces generated by NatOcc using color transfer, image harmonization, and super-resolution techniques. The last two rows show occluded faces generated using RandOcc by overlaying random shapes with random textures and transparency.}
  \label{fig:teaser}
  \vspace{-4mm}

\end{figure}

Supervised training of occlusion-aware face segmentation requires a large amount of data. However, existing real-world occluded face datasets such as~\cite{dataset_cofw,dataset_CelebAMask-HQ,dataset_part} are not suitable for this purpose. They are either low in quantity, low resolution or inaccurately labeled. As data collection and annotation is very time-consuming and labor-intensive, previous studies~\cite{nirkin2017face,yin2021segmentationreconstructionguided,saito2016realtime,dataset_redondo2020extended} created their synthetic datasets for training by overlaying commonly wearable objects or hand on existing face datasets with basic augmentation techniques. They often replaced the occluders' texture and color to enrich the occlusions' diversity. 
 
\begin{figure}[t]
  \centering
  \includegraphics[width=1\linewidth]{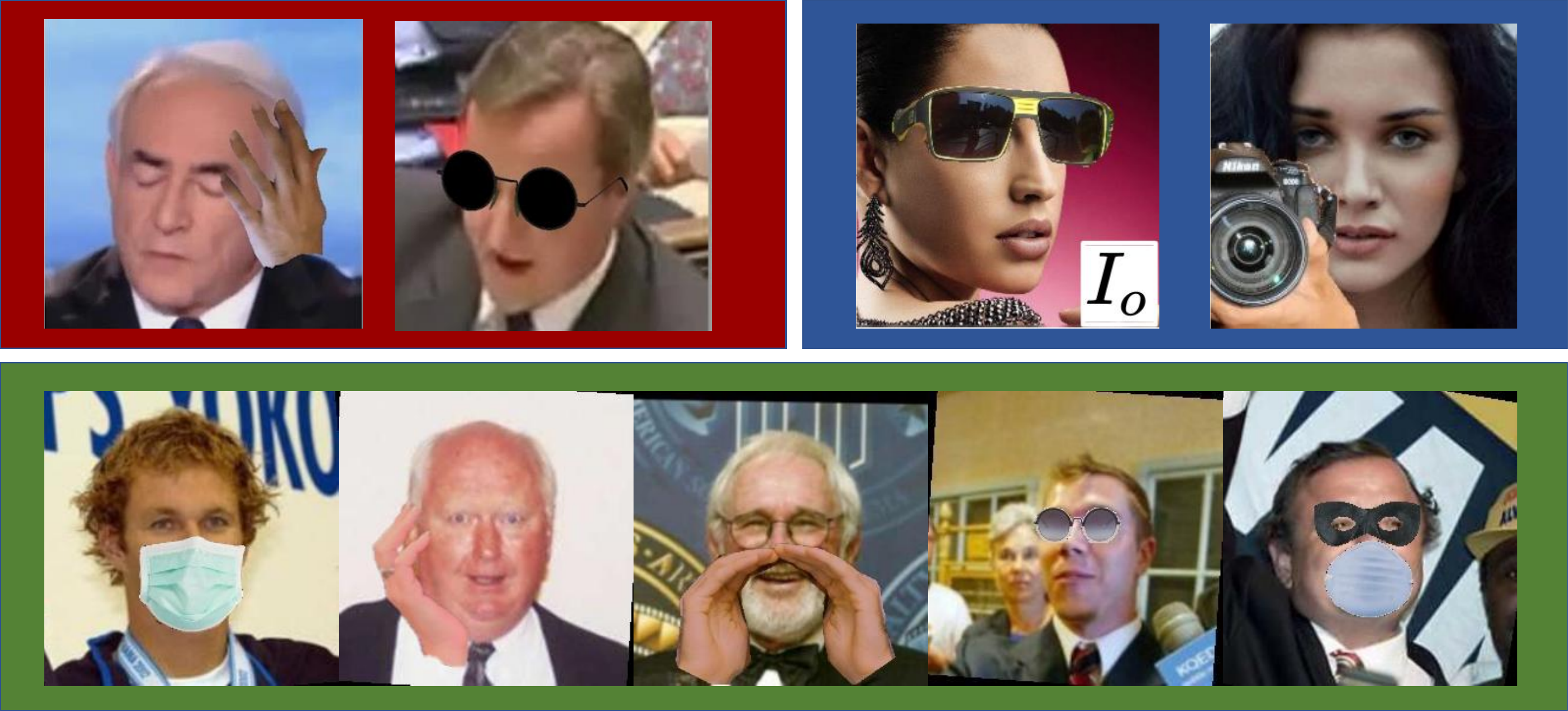}
  \vspace{-5mm}

  \caption{Examples of synthetic occluded face images taken from previous studies. The red box shows images from~\cite{nirkin2017face}, whereas the blue box shows images from~\cite{yin2021segmentationreconstructionguided}, and the green box shows images from~\cite{dataset_redondo2020extended}. Specific types of occluders (\emph{e.g.}, sunglasses, face masks, and hands) were overlaid at a specific location, and the color and texture of occluders often looked unnatural.}
  \label{fig:other_work}
  \vspace{-5mm}
\end{figure}

Nevertheless, synthetic data generation often neglects the naturalistic aspects of the data. The color, texture, and edges of the occluders are visually unnatural, as shown in Figure~\ref{fig:other_work}. In addition, these synthetic data generation efforts~\cite{dataset_redondo2020extended,yin2021segmentationreconstructionguided} usually only work for specific use cases and cannot scale to other applications. Specific types of occluders (\emph{e.g.}, sunglasses, and mask) were overlaid at a fixed location, as shown in Figure~\ref{fig:other_work}. The question then arises whether more general synthetic data generation could perform as well, if not better than, or at least comparable with these specific methods. Moreover, the commonly used validation set, Caltech Occluded Faces in the Wild (COFW) dataset \cite{dataset_cofw} might not be a suitable benchmark since not all the faces are occluded due to different occlusion definitions. Thus, a new standard benchmark is required.

In this study, we proposed two data generation techniques: Naturalistic Occlusion Generation (NatOcc), an effective method to produce naturalistic synthetic occluded face from CelebAMask-HQ~\cite{dataset_CelebAMask-HQ} using various techniques such as color transfer, image harmonization, and super-resolution; and Random Occlusion Generation (RandOcc), a more general method by overlaying random shapes with random transparency and textures from the Describable Textures Dataset (DTD)~\cite{dataset_dtd}. To evaluate the effectiveness and robustness of our synthetic datasets, they were compared with a real-world occluded face dataset.
The real-world occluded face dataset was prepared by manually correcting approximately 4,000 incorrectly labels from CelebAMask-HQ \cite{dataset_CelebAMask-HQ}. The dataset is further split into occluded and non-occluded categories.
Our synthetic datasets were generated from the non-occluded images. Some examples of the synthetic data generated by NatOcc and RandOcc are shown in Figure~\ref{fig:teaser}.
Both data generation methods performed at a level or even better than real-world occluded face dataset.
Besides, our methods were proved to be effective and robust to handle unseen occlusions and faces in the wild (see Section~\ref{sec:result}).
Furthermore, to facilitate model evaluation, we present two high-resolution real-world occluded face datasets collected from Pexels and Unsplash with fine-grained manually annotated masks: RealOcc, which consists of 550 aligned and cropped face images; and RealOcc-Wild, which comprises 270 occluded face images in the wild for robustness test.

Our contributions are summarized as follows.
\textbf{1)} We propose two synthetic occluded data generation techniques, NatOcc to produce naturalistic synthetic occluded faces, and RandOcc, a general synthetic occluded data generation method.
\textbf{2)} We provide manually corrected annotation masks and new categories (occluded and non-occluded) for widely used CelebAMask-HQ~\cite{dataset_CelebAMask-HQ}.
\textbf{3)} To facilitate model evaluation, we contribute two real-world occluded face datasets with manually annotated masks, RealOcc (aligned and cropped) and RealOcc-Wild (in the wild).
\textbf{4)} We further benchmark several representative baselines~\cite{zhao2017pspnet,deeplabv3plus2018,xie2021segformer} and present a comprehensive analysis, verifying the effectiveness of our method (even for unseen occlusions) and providing insights for future exploration.

\section{Related Work}
\label{sec:related_work}

\noindent \textbf{Real-World Occluded Face Datasets.}
Existing real-world occluded face datasets~\cite{dataset_cofw,dataset_LFWTech,dataset_yang2016wider,dataset_CelebAMask-HQ} have the following shortcomings: they are either low in quantity, low resolution, incorrectly labeled, or lack of annotated masks. To our knowledge, there are only a limited number of real-world occluded face datasets \cite{dataset_CelebAMask-HQ,dataset_cofw} that cover various occlusions (\emph{e.g.}, sunglasses, hats, foods, and hands). 
%
%
For instance, Real World Occluded Faces (ROF) \cite{dataset_erakiotan2021recognizing} consists of 1,686 sunglasses-occluded faces and 678 mask-occluded faces. The datasets that are specifically targeted for face mask occlusion are Interactive Systems Labs (ISL) Unconstrained Face Mask Dataset (ISL-UFMD) \cite{dataset_eyiokur2021computer} with 10,618 face images, and the Face Mask Label Dataset (FMLD) \cite{dataset_app11052070} with 63,072 face images. Additionally, Specs on Faces (SoF) \cite{dataset_afifi2017afif4} consists of 42,592 glasses-occluded face images and Interactive Systems Labs (ISL) Unconstrained Face Hand Interaction Dataset (ISL-UFHD) \cite{dataset_eyiokur2021computer} consists of 10,004 hand-occluded images. Many of these datasets do not have annotated masks. 

The only large-scale, high-quality dataset with masks is CelebAMask-HQ \cite{dataset_CelebAMask-HQ}. This dataset contains 30,000 face images (1,024$\times$1,024), each containing masks of different parts of the face and accessories such as sunglasses, hats, and earrings. Out of these 30,000 images, approximately 4,000 images are occluded faces. However, the occluded faces suffer incorrect annotations as some objects (\emph{e.g.}, microphones, food, and hands) that overlap the face area are annotated as part of the face rather than the background. The COFW dataset \cite{dataset_cofw} used to have 1,007 annotated masks, but the full data is no longer publicly available. A previous study \cite{dataset_Ghiasi2015UsingST} has provided the 500 annotated masks for the training set of the COFW \cite{dataset_cofw} dataset. There are a few shortcomings with this dataset. First, the quantity and resolution of images in the training set are too low, and the masks of the test partition \cite{dataset_cofw}, which were used as a benchmark in different studies, are not available. Lastly, Part Labels \cite{dataset_part}, which is a subset from LFW \cite{dataset_LFWTech}, contains only 2,927 images with annotated masks. However, these masks are coarse, not accurately annotated and lacked variety in occlusion. Hence, it is not suitable to be both training and test set.

\noindent \textbf{Synthetic Occluded Face Datasets.}
Numerous approaches have been proposed to generate synthetic occluded faces. Each method has its own considerations, leading to varying numbers of classes and definitions of occlusions. Many studies such as \cite{saito2016realtime,nirkin2017face,yin2021segmentationreconstructionguided,dataset_redondo2020extended} generated their synthetic datasets by overlaying common occlusions such as sunglasses, masks, hands onto faces. However, most of them are built on a low-resolution dataset, and the synthetic occluded faces do not look natural due to the occluders' color, texture, and edges. Besides, these data generation methods usually overlaid specific types of occluders at specific locations and orientations. Since most of their code or datasets are not made public, it is challenging to reproduce, cross-check, and improve on previous data generation methods. The existing occlusion augmentation techniques such as~\cite{aug_Zhong_Zheng_Kang_Li_Yang_2020,aug_devries2017improved} are not used in previous studies. Instead, they produce unnatural occlusions by covering a region of the image with rectangles. 

Extended Labeled Faces in-the-Wild (ELFW) \cite{dataset_redondo2020extended} has 3,754 labeled faces (250$\times$250), which is an extension from Part Labels \cite{dataset_part}. The original Part Labels \cite{dataset_part} has many incorrect annotations and lacks occluded faces. Thus, the contribution of \cite{dataset_redondo2020extended} went beyond simply improving the masks of the original dataset by adding new categories (\emph{e.g.}, sunglasses, head-wearables and face masks), but also increased the occluded face images by overlaying objects and hands over the original images. They followed the approach described in \cite{dataset_nojavanasghari2017hand2face} to perform occlusion augmentation of hands from both Hand2Face \cite{dataset_nojavanasghari2017hand2face} and HandOverFace \cite{dataset_khan2018analysis}.  
Although the synthetic dataset is not provided, their code to perform occlusion augmentation is released. The main issue with this dataset is that the resolution and quantity are low. 

The more recent work \cite{yin2021segmentationreconstructionguided} built their synthetically occluded training dataset based on CelebAMask-HQ \cite{dataset_CelebAMask-HQ} where they occluded the dataset with 300 different occluders (\emph{e.g.}, masks, scarfs) and replaced the texture on the original occluders from 800 textures to create more variation. A similar concurrent work to our study is FaceOcc \cite{dataset_faceocc}, where they fixed the masks of CelebAMask-HQ \cite{dataset_CelebAMask-HQ} and produced a synthetic occluded face dataset with it. However, they have different definitions of occlusions. At the time of this study, they had yet to release their code or dataset. Therefore, further comparison was not included.


\section{Methodology}

\subsection{Dataset Preparation }
This section shows our rationale in dataset selection and methods to prepare different datasets. First of all, it is essential to set some crucial definitions as each previous study had a different interpretation. Our study has two classes, background and face, the grey or ambiguous area of which is treated as that category (see Table \ref{tab:occdef}) at the time of this study.

\begin{table}[H]
   \vspace{-2mm}

    \renewcommand{\arraystretch}{1} 
  \centering
\caption{Definition of classes in our dataset.}
  \vspace{-3mm}
   \resizebox{0.9\linewidth}{!}
    {
  \begin{tabular}{p{0.23\linewidth}   p{0.67\linewidth}}
    \toprule
    Class & Definition \\
    \midrule
    Face & The skin of the head includes eyes, nose, mouth but excluding ears.   \\

    Face\newline(Grey Area) & Tattoo, shadow, moustache, and beard overlapped with face, as well as skin of the bald person are considered as part of the face.  \\
                    
    Background (Occlusions) & Non-face area and any objects such as sunglasses, shirt, hair, microphone, hands that are physically covering (overlap) the face. Words from magazines or copyright labels fall into this category as well.   \\
    Background (Grey Area) & Transparent/Translucent glasses\\
    \bottomrule
  \end{tabular}
  }

  \label{tab:occdef}
    \vspace{-5mm}

\end{table}

\begin{figure}[h]
\centering
\includegraphics[width=1\linewidth]{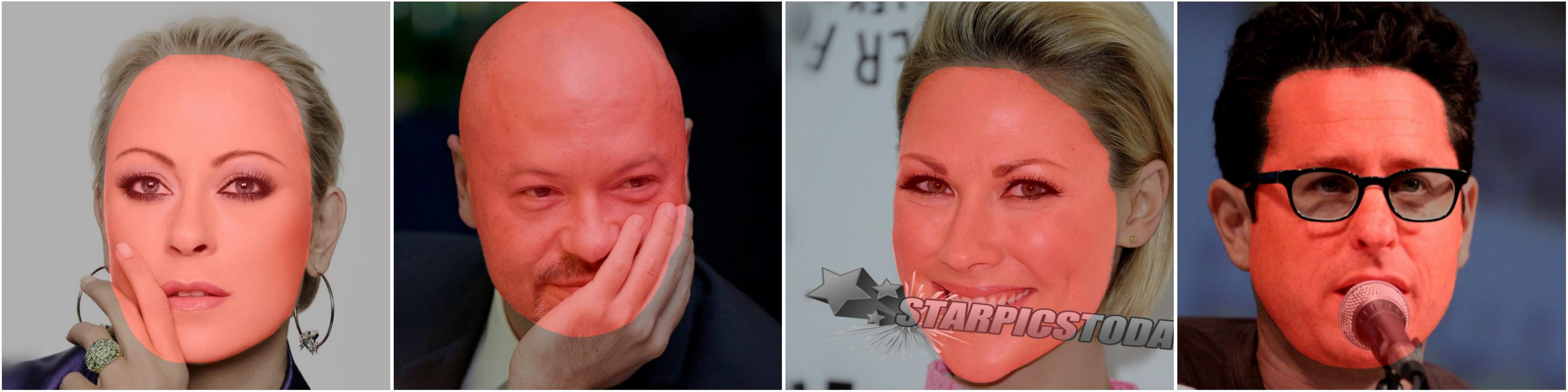}
\vspace{-6mm}
\caption{Examples of the wrongly annotated masks in CelebAMask-HQ \cite{dataset_CelebAMask-HQ} as occlusions are treated as part of the face.} 
\label{fig:celeb_occlusion}
\vspace{-3mm}
\end{figure}

\noindent \textbf{Face Dataset.} CelebAMask-HQ \cite{dataset_CelebAMask-HQ} is a suitable and relevant face dataset since it contains 30,000 high-quality images with refined masks. The initial face masks were obtained by subtracting the skin masks with the masks of sunglasses, neck, and head in the dataset. After that, we manually corrected the wrong annotated masks (see Figure~\ref{fig:celeb_occlusion}) such as hands and microphone using an online annotation platform, Segments.ai\footnotemark. Next, the dataset was split into occluded and non-occluded categories. The occluded category refers to faces that are overlapped or intersected with any objects. Most of the corrupted and cartoon images were excluded from the dataset. A subset of images (716 images) made of 80 identities was extracted from the non-occluded category to act as the test set. The partition is summarized in Table \ref{tab:celeb_part}.\footnotetext{\url{https://segments.ai/}}

\begin{table}[H]
  \vspace{-3mm}
  \renewcommand{\arraystretch}{1} 
  \centering
  \captionsetup{justification=centering}
  \caption{Summary of our partition of CelebAMask-HQ \cite{dataset_CelebAMask-HQ}.}
  \vspace{-2mm}
   \resizebox{0.9\linewidth}{!}
    {
  \begin{tabular}{p{0.6\linewidth}   p{0.3\linewidth}}
    \toprule
    Category & Quantity \\
    \midrule
    CelebAMask-HQ-WO (Train) & 24603\\
    CelebAMask-HQ-WO (Test) & 716 \\
    CelebAMask-HQ-O &   4597 \\
    Excluded Images & 86 \\
    \bottomrule
  \end{tabular}
  }
  \vspace{-3mm}

  \caption*{WO - Without occlusion \hspace{10pt}  O - Occluded}
  \label{tab:celeb_part}
  \vspace{-2mm}
\end{table}

\begin{figure}[t]
\centering
\includegraphics[width=1\linewidth]{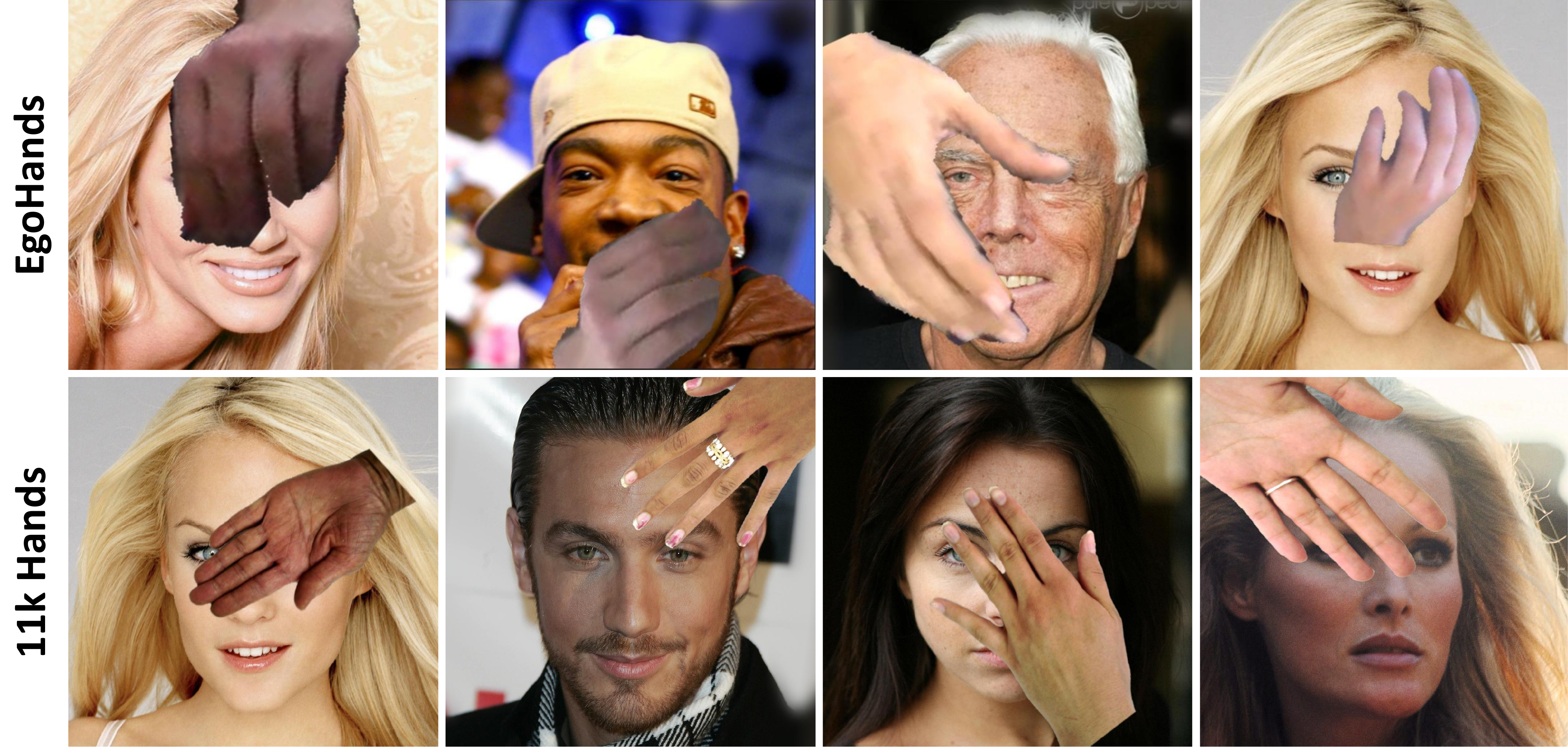}
\vspace{-4mm}
\caption{Comparison of the images from CelebAMask-HQ~\cite{dataset_CelebAMask-HQ} overlaid with different hands datasets. The hands from 11k Hands~\cite{dataset_afifi201911kHands} have finer details and higher resolution than EgoHands~\cite{dataset_bambach2015lending}.}
\label{fig:celeb_with_hands}

\vspace{-2mm}
\end{figure}

\noindent \textbf{Occluders Dataset.} 
Unlike other studies \cite{saito2016realtime,nirkin2017face,yin2021segmentationreconstructionguided,dataset_redondo2020extended,dataset_faceocc} that use very specific occluders such as sunglasses and face mask, we extracted 128 common objects across 20 categories (\emph{e.g.}, food, bottles, cellphones, and cups) from the Microsoft Common Objects in Context (COCO) dataset~\cite{dataset_lin2015microsoft} with COCOAPI \cite{dataset_lin2015microsoft}. The original resolutions of the COCO objects were low, so we performed super-resolution ($\times$4) of the original images with GLEAN \cite{chan2021glean}. In contrast to other occlusions, hands have similar color to faces, making them harder to detect as occlusion. There are several existing hand datasets \cite{dataset_bambach2015lending,dataset_khan2018analysis}. Initially, we used the 200 hands of EgoHands \cite{dataset_bambach2015lending} that were sampled by \cite{nirkin2017face}. However, the image resolution of the hands is low and blurry. The edges and details of the hands are not preserved after resizing and overlaid onto CelebAMask-HQ \cite{dataset_CelebAMask-HQ} images, as shown in Figure \ref{fig:celeb_with_hands}. Therefore, we sampled 200 images from another hand dataset, 11k Hands \cite{dataset_afifi201911kHands}, which has a higher resolution of 1,600$\times$1,200. Due to the lack of fine masks, we manually annotated the 200 hands images. The only drawback of this hands dataset is the lack of different postures, but it is good enough for our study. The comparison of the hands overlaid by both datasets is shown in Figure \ref{fig:celeb_with_hands}, the hands from 11k hands \cite{dataset_afifi201911kHands} have finer details and higher resolution than EgoHands \cite{dataset_bambach2015lending}.

\begin{figure}[t]
\centering
\includegraphics[width=1\linewidth]{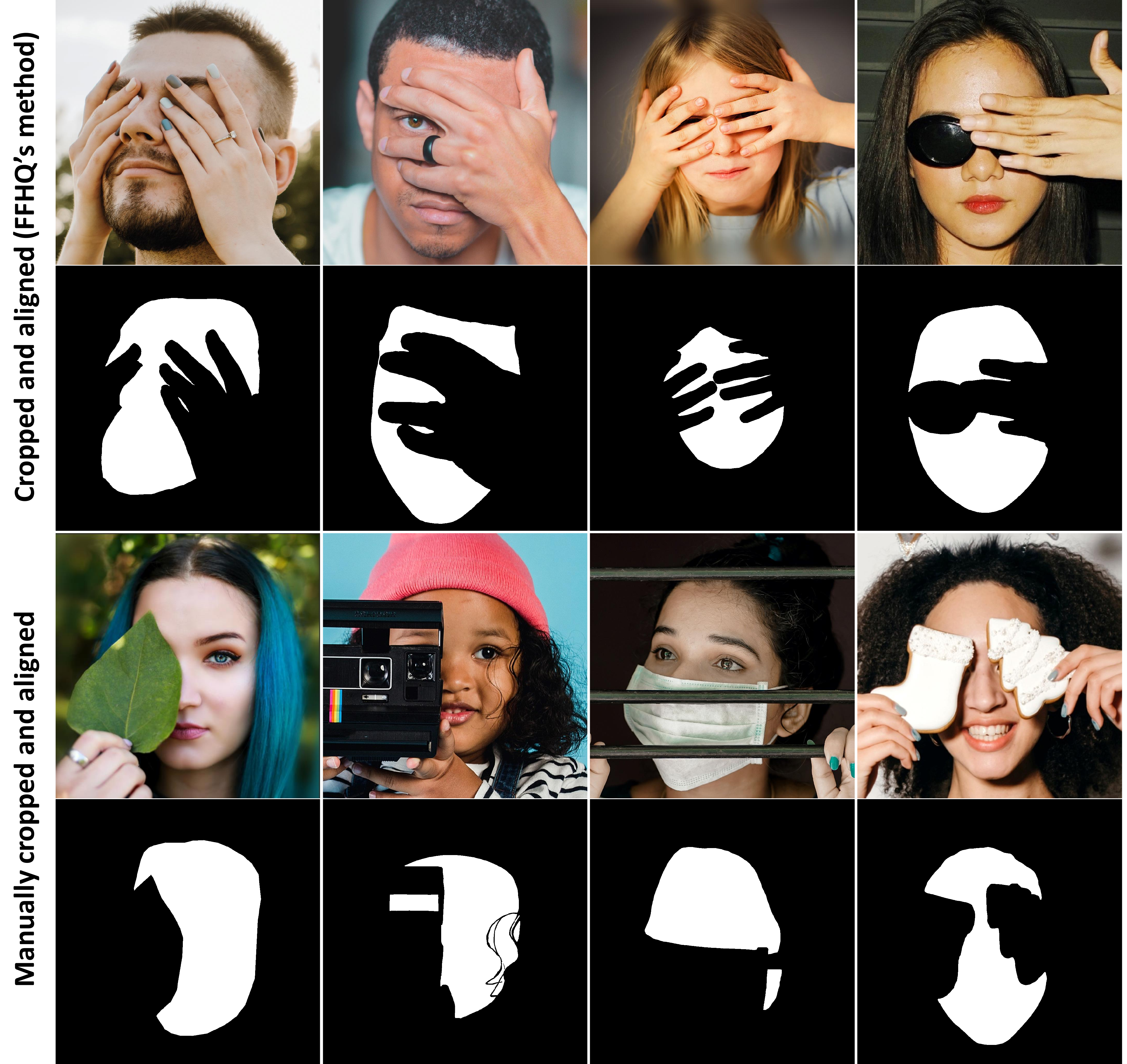}
\vspace{-4mm}
\caption{Examples of aligned and cropped occluded face images in RealOcc with different methods.} 
\label{fig:RealOcc}
\vspace{-1mm}

\end{figure}

\noindent \textbf{Validation Set.} To facilitate model evaluation, we introduce a new validation set, RealOcc consisting of 550 high resolution (1,024$\times$1,024) occluded face images from websites such as Pexels and Unsplash, covering different occlusions (\emph{e.g.}, hands, masks, food and sunglasses). The face was aligned and cropped as shown in Figure \ref{fig:RealOcc} using the same method used in FFHQ \cite{dataset_karras2019stylebased}. However, due to occlusions, approximately 265 occluded face images failed to be detected by dlib \cite{dlib09}. With reference to the aligned and cropped images, we manually aligned and cropped the 265 face images. To speed up the labeling process, coarse masks were produced with the models that are trained with our synthetic dataset, followed by correcting the mask using LabelMe \cite{Wada_Labelme_Image_Polygonal} and Segments.ai. We introduced another validation dataset for model robustness test, RealOcc-Wild, consisting of 270 occluded faces in the wild (without cropping and alignment) that are mainly made of the images failed to be detected by dlib \cite{dlib09}. Due to our definition of occlusion (\emph{e.g.}, transparent glasses are occlusion, and mustache overlapping faces is not occlusion), the masks in the COFW \cite{dataset_cofw} dataset were manually modified.

\subsection{Naturalistic Occlusion Generation (NatOcc)}
\label{sec:datasetgen}

Due to the lack of real-world, large-scale occluded face datasets, previous studies \cite{nirkin2017face,yin2021segmentationreconstructionguided,saito2016realtime,dataset_redondo2020extended,dataset_faceocc} have proposed different synthetic occluded face generation techniques. In this work, we propose a NatOcc method to produce high-quality naturalistic occluded face images from CelebAMask-HQ-WO (Train) (see Table \ref{tab:celeb_part}).

\begin{algorithm}[h]
    \caption{Color transfer via Sliced Optimal Transport (SOT) \cite{bonneel19SPOT} with custom preprocess}
    \label{algo:sot}
    $sQuantity$ $\leftarrow$ $source$ $<$ $lowerThresh$\;
    $tQuantity$ $\leftarrow$ $target$ $== 0$ \;
    $blackRatio$ $\leftarrow$ $sQuantity / tQuantity$\;

    \If{$blackRatio>1$}
      {
       $mean(rgb)$ $\leftarrow$  Mean of non-black $source$ pixels in each channel \; 
        Replace $blackRatio-1$ ratio of black pixels in $source$ with $mean (rgb)$\;
      }
    Clipped pixels value of $source$ to $upperThresh$\;
    ColorTransferViaSOT($source$,$target$)\;
\end{algorithm}

\begin{figure}[t]
    \centering
    \includegraphics[width=0.9\linewidth]{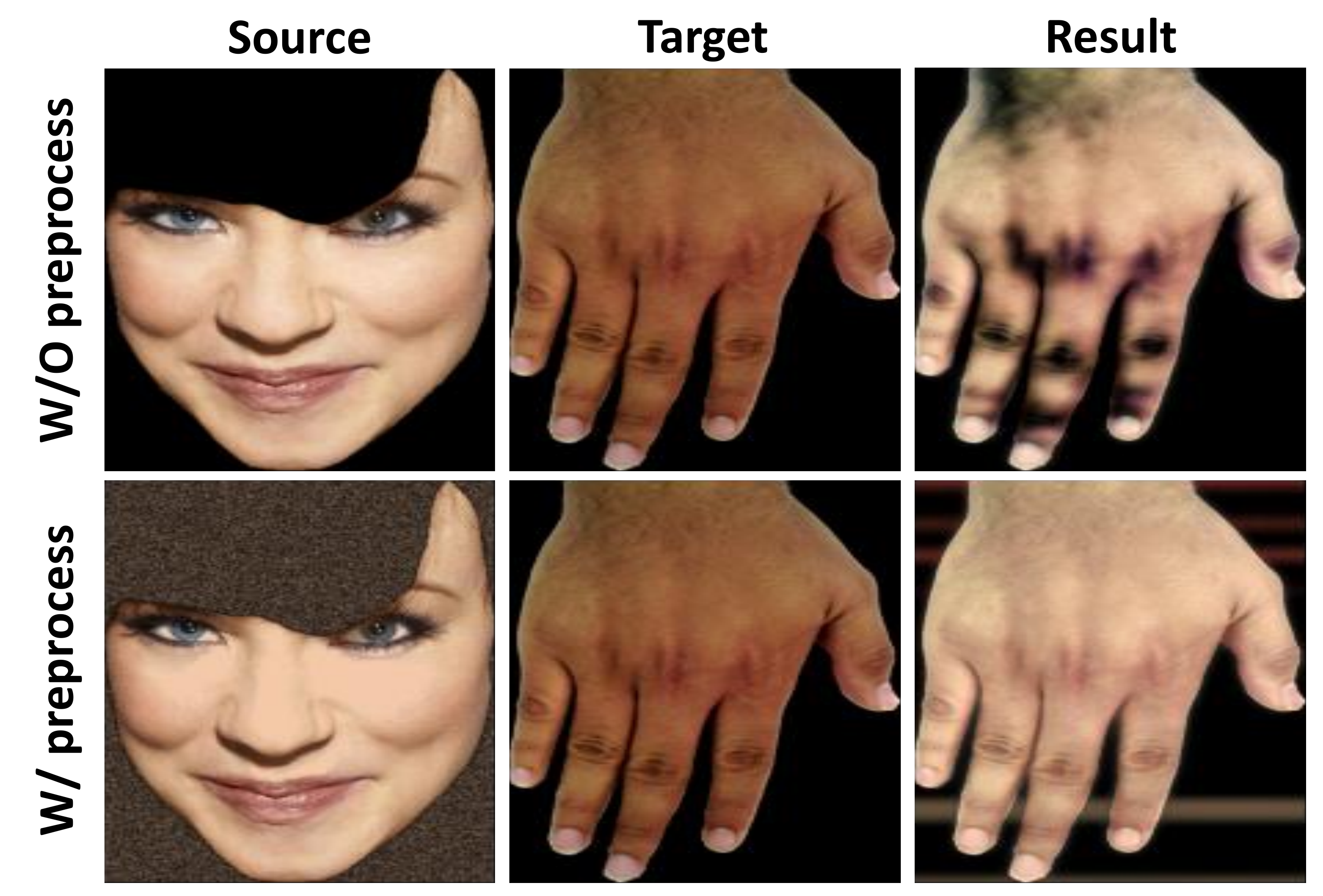}
    \vspace{-1mm}
    \caption{Comparison of the color transfer via Sliced Optimal Transport (SOT) \cite{bonneel19SPOT} with or without custom preprocess. The preprocess mitigates the issue of black pixels imbalance between the source and the target images, thus improving the quality of the color transfer.} 
    \label{fig:sot}
    \vspace{-2mm}

\end{figure}

\begin{figure}[t]
    \centering
    \includegraphics[width=1\linewidth]{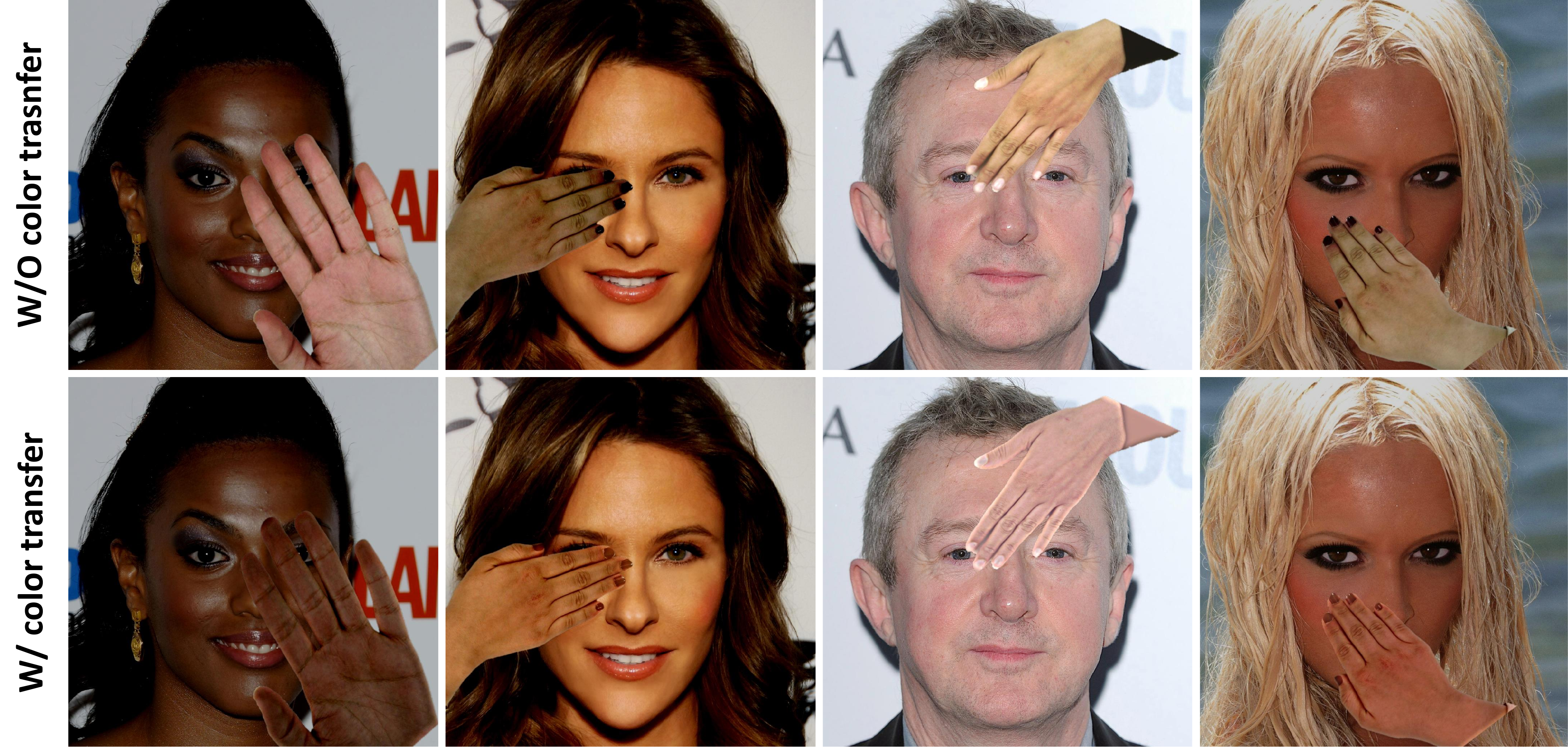}
    \vspace{-4mm}

    \caption{The effects of color transfer via SOT~\cite{bonneel19SPOT} on 11k Hands~\cite{dataset_afifi201911kHands} with CelebAMask-HQ-WO (Train) images.} 
    \label{fig:sot_example}

\end{figure}

\noindent \textbf{Color Transfer.} In reality, most hand occlusions have a similar color to the face. To simulate this scenario, colors from the face were transferred to the hands using color transfer via Sliced Optimal Transport (SOT) \cite{bonneel19SPOT}. The size of the source (face) and target (hand) images must be the same.
If the quantity of the black pixels of the source (face) is larger than those of the target (hand), some area of the hands will appear black due to black pixels imbalance, as shown in Figure~\ref{fig:sot}. On the other hand, if a part of the pixels of the source (face) are too bright, the hands will look unnatural as well.
Therefore, preprocess is necessary to address this issue. To resolve this issue, a ratio of black pixels at the source (face) were replaced with the average of each RGB channel of the non-black pixels at the source (face) correspondingly, and the pixels values of the source (face) image were clipped at a certain threshold.
The steps of the preprocess before the color transfer is shown in Algorithm~\ref{algo:sot}. The comparison of the color transfer with and without preprocess is shown in Figure~\ref{fig:sot}. This simple preprocess can solve most of the cases. However, future work can look into color transfer via Sliced Partial Optimal Transport (SPOT) \cite{bonneel19SPOT} that might handle the above issues better. To speed up the long process of this method, we have utilized GPU and multiprocessing. Samples of the synthetic images with or without color transfer can be found in Figure~\ref{fig:sot_example}.
 
\hypertarget{Augmentation}
\noindent \textbf{Augmentation.} Affine augmentation, image compression, random brightness and contrast were applied to both faces and occluders using Albumentations \cite{info11020125}. Besides, occluders were randomly resized to 0.5-1 of the face size. Moreover, the edges of occluders were carefully considered to produce a more naturalistic composite image. This was done by applying Gaussian blur to the mask of the occluders before alpha blending, a method of overlaying a foreground image over a background image. After alpha blending, the intersection of the occluders and the faces in the composite image was blurred again. The occluders were randomly positioned around the faces.

\noindent \textbf{Additional Augmentation (Hand Only).} Besides color transfer, hands were positioned so that the fingers always point to the face, followed by a small random rotation to simulate hands in real-life scenarios.

\begin{figure}[t]
    \centering
    \includegraphics[width=1\linewidth]{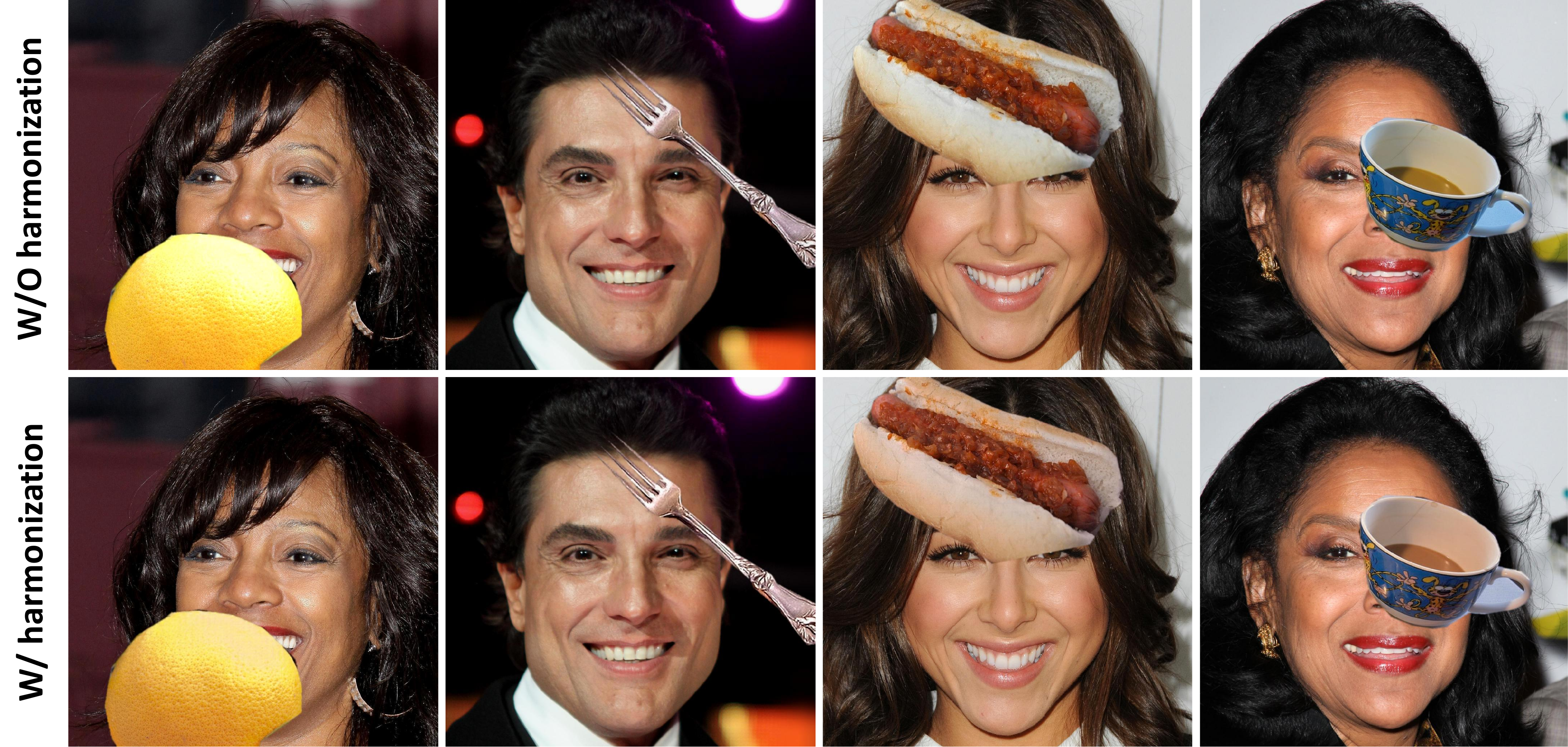}
    \vspace{-4mm}

    \caption{The effects of image harmonization on the occluders using RainNet \cite{ling2021region}. The images after image harmonization look more natural. } 
    \label{fig:colour_harmonization}
    \vspace{-2mm}

\end{figure}

\noindent \textbf{Image Harmonization.} To further improve the naturalistic aspect of synthesized images, we applied RainNet \cite{ling2021region} to harmonize the foreground (occluder) to match the background (face image). COCO \cite{dataset_lin2015microsoft} occluders look more natural after image harmonization (see Figure~\ref{fig:colour_harmonization}). However, it was not the same case for the hand occluder as the color of the hand was changed after image harmonization, defeating the purpose of color transfer. Thus, image harmonization was not applied to the hand occluder.
\subsection{Random Occlusion Generation (RandOcc)}
\label{sec:RandOcc}

To overlay hands onto the faces, studies such as  \cite{dataset_redondo2020extended,dataset_nojavanasghari2017hand2face} find the matching target face images with the same posture as the source face image, followed by overlaying the source image's hands onto the target face image. Occluders such as sunglasses and face masks were overlaid onto eyes and mouth, respectively. Such a synthetic dataset generation method cannot be applied to most applications. Thus, we propose a more general occlusion method, Random Occlusion Generation (RandOcc), which creates synthetic occluded data with minimal effort. The performance and robustness of this RandOcc dataset will be compared with the real occluded dataset and our NatOcc dataset. 

\noindent \textbf{Occlusion Augmentation.} RandOcc overlaid random shape with random transparency and texture from Describable Textures Dataset (DTD) \cite{dataset_dtd}, which consists of 5,640 images from 47 categories. The same augmentation process (see Section~\hyperlink{Augmentation}{3.2}) was applied in RandOcc. 

\begin{figure}[t]

\centering
\includegraphics[width=1\linewidth]{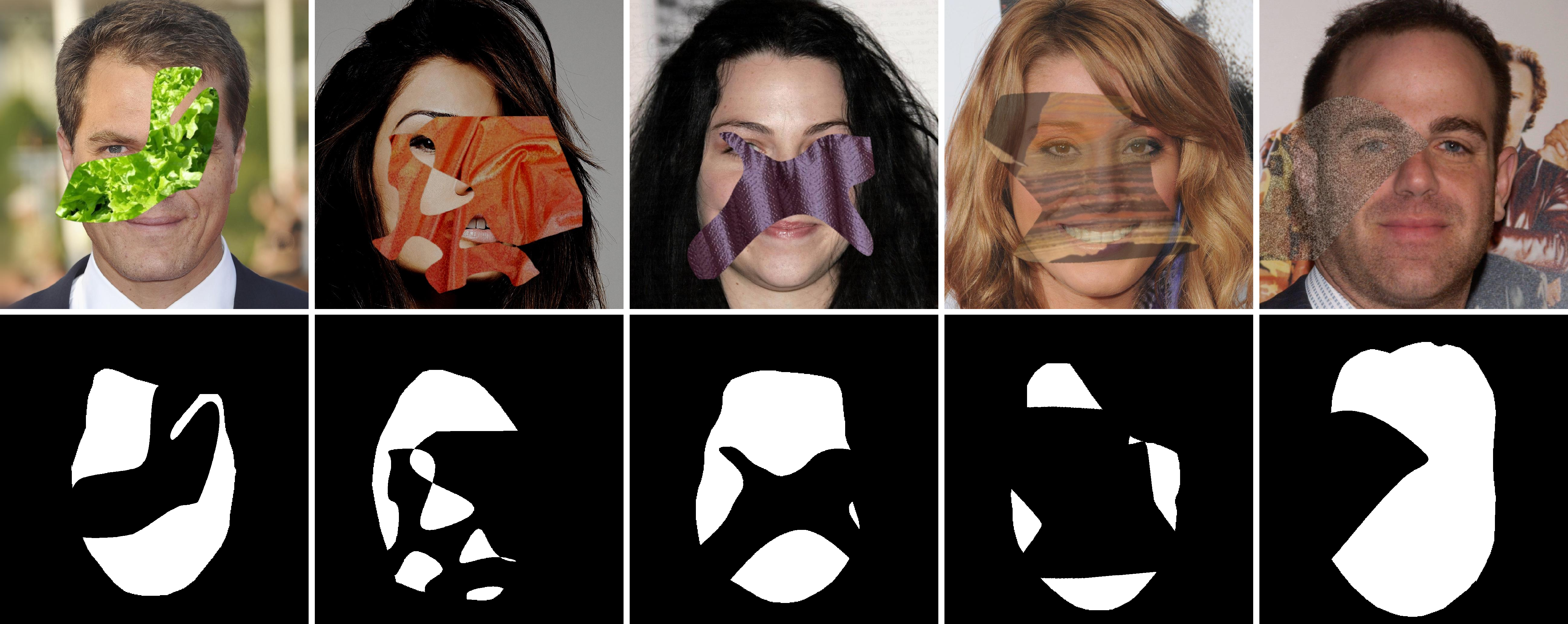}
\vspace{-4mm}
\caption{Examples of CelebAMask-HQ-WO (Train) occluded with random shape and texture from DTD \cite{dataset_dtd}. 30 percent of the occluders are slightly transparent.} 
\label{fig:RandOcc}
    \vspace{-1mm}

\end{figure}

\noindent \textbf{Transparent/Translucent Object Simulation.} \label{sec:simulate_transparency} To simulate transparent or translucent objects, the alpha mask of alpha blending was randomly set between 0.5 to 0.8. This is applied to approximately 30 percent of the dataset. Examples of the RandOcc dataset can be found in Figure~\ref{fig:RandOcc}.

\section{Experiments}

\subsection{Settings}

\noindent We provide quantitative and qualitative results on different variants of our dataset. The training was carried out with two CNNs, PSPNet \cite{zhao2017pspnet} and DeepLabv3+~\cite{deeplabv3plus2018} with pre-trained ResNet-101 backbone \cite{he2016deep}, and a Vision Transformer, SegFormer \cite{xie2021segformer} with pre-trained MIT-B5 backbone. The trained models were tested on two datasets, COFW (train set) \cite{dataset_cofw}, and RealOcc-Wild to compare the robustness of each model. The test results of CelebAMask-HQ-WO (Test) are provided in the \emph{Appendix}.


\noindent \textbf{Implementation Details.} All experiments were carried out using MMSegmentation~\cite{mmseg2020}. Every model was trained on 4 Tesla V100 GPUs, with an input image size of 512$\times$512 and a batch size of 8 for 30k iterations. The optimizer for PSPNet \cite{zhao2017pspnet} and DeepLabv3+ \cite{deeplabv3plus2018} is SGD with learning rate of 0.01, momentum of 0.9 and decay rate of 0.0005. The optimizer for SegFormer \cite{xie2021segformer} is AdamW with a learning rate of 6e-05 and weight decay rate of 0.01. Evaluations take place every 400 iterations on the RealOcc dataset. In all the experiments, images were resized to 512$\times$512. Basic data augmentations, such as random horizontal flips, 30 degrees of random rotation, and photometric distortion were applied. The loss function for all the experiments is the binary cross-entropy loss with online hard example mining (OHEM)~\cite{ohem}.

\noindent \textbf{Datasets.} The definition of training datasets is shown in Table \ref{tab:data_def}. Different combinations of training datasets can be found in Table \ref{tab:result}. All the synthetic datasets were generated using C-WO, \emph{i.e.}, CelebAMask-HQ-WO (Train) (see Table \ref{tab:celeb_part}). The validation sets are the RealOcc, while the test sets are COFW (training set) \cite{dataset_cofw} and RealOcc-Wild.

\begin{table*}[ht]

    \renewcommand{\arraystretch}{1} 
  \centering
    \caption{Definition of datasets in our experiments.}
  \vspace{-2mm}

   \resizebox{0.9\textwidth}{!}
    {
  \begin{tabular}{p{0.2\textwidth}   p{0.7\textwidth}}
    \toprule
    Class & Definition \\
    \midrule
    C-Original & CelebAMask-HQ-WO (Train) and CelebAMask-HQ-O with original masks.\\
    C-CM & CelebAMask-HQ-WO (Train) and CelebAMask-HQ-O with corrected masks.\\
    C-WO &  CelebAMask-HQ-WO (Train).\\
 
    C-WO-NatOcc &  One set of hand-occluded (without color transfer) face dataset  \newline and one set of COCO-occluded face dataset generated by NatOcc with C-WO. \\
    C-WO-NatOcc-SOT &  One set of hand-occluded (with color transfer) face dataset  \newline and one set of COCO-occluded face dataset generated by NatOcc with C-WO. \\
    C-WO-RandOcc &  Two sets of occluded face dataset generated by RandOcc with C-WO. \\
    C-WO-Mix &  Half set of C-WO-RandOcc \newline and one set of C-WO-NatOcc.  \\
    \bottomrule
  \end{tabular}
  }

  \label{tab:data_def}

\end{table*}

\noindent \textbf{Evaluation Metrics.} We evaluate the model performance by comparing mIoU, \emph{i.e.}, the mean Intersection over Union, across all the classes on the validation dataset.

\label{sec:result}

\begin{table*}[ht]
    \centering
  \caption{ \textbf{Overall Performance:} Results of PSPNet \cite{zhao2017pspnet}, DeepLabv3+ \cite{deeplabv3plus2018} and SegFormer \cite{xie2021segformer} with different combination of datasets. The best results for each validation set are marked in bold. The metrics are mIoU (higher is better).}
    \vspace{-2mm}

    \resizebox{1\textwidth}{!}
    {
    \begin{tabular}{lcccccccccccccc}
    \toprule
    & Quantity &&\multicolumn{3}{c}{\textbf{RealOcc} (mIoU)} &
    &\multicolumn{3}{c}{\textbf{COFW (Train)} (mIoU)} &
    &\multicolumn{3}{c}{\textbf{RealOcc-Wild }(mIoU)}\\
    \cmidrule{4-6} \cmidrule{8-10} \cmidrule{12-14} 
    & & &  PSPNet & DeepLabv3+  & SegFormer    
       & & PSPNet & DeepLabv3+  & SegFormer   
       & & PSPNet & DeepLabv3+  & SegFormer    \\

    \midrule
     C-Original & 29,200                       && 89.52 & 88.13 & 88.33             && 89.64&88.62&91.36  && 85.21 &82.05& 85.24\\
     C-CM & 29,200                    && 96.15 & 96.13 & 97.42            && 91.82&92.77&\textbf{94.87}  && 91.33 & 91.01 & 95.16 \\
     C-WO  & 24,602                           && 89.38 & 89.01& 91.36               && 89.53&88.97&92.24 && 83.86 & 84.14 & 86.72 \\               
     C-WO + C-WO-NatOcc & 24,602 + 49,204        && 96.65 & 96.51 & 97.30                  && 90.71&91.21& 94.30 && 91.34 & 91.70 & 94.17 \\ 
     C-WO + C-WO-NatOcc-SOT & 24,602 + 49,204    && 96.35 & 96.59 & 97.18                  && \textbf{92.32}&91.74&93.55 && \textbf{93.26} & 92.69 & 94.27 \\

     C-WO + C-WO-RandOcc & 24,602 + 49,204       && 95.09 & 95.21 & 96.53                       && 90.82&91.35&93.14        && 89.54 & 89.68 & 92.84\\
     C-WO + C-WO-Mix & 24,602 + 73,806  && 96.55 & 96.66 & 97.37                 &&    90.99&91.20 & 93.74     &&  92.14 & 91.84 & 94.40\\
      C-CM + C-WO-NatOcc & 29,200 + 49,204  && \textbf{97.28} & \textbf{97.33} & 97.95 && 91.61&92.66&94.86 && 92.13 & \textbf{93.81} & \textbf{95.43}\\ 
      C-CM + C-WO-NatOcc-SOT & 29,200 + 49,204  && 97.17 & 97.29 & \textbf{98.02}      && 92.07&\textbf{92.91}&94.60 && 92.84 & 93.73 & 94.53\\

    \bottomrule
    \end{tabular}
    }


  \label{tab:result}
  

\end{table*}

\subsection{Results and Analysis}
We evaluate our two data generation methods with different approaches. NatOcc focuses on producing naturalistic occluded faces, while RandOcc produces occlusions with a general approach.

\begin{figure*}[t]
\centering
\includegraphics[width=1\linewidth]{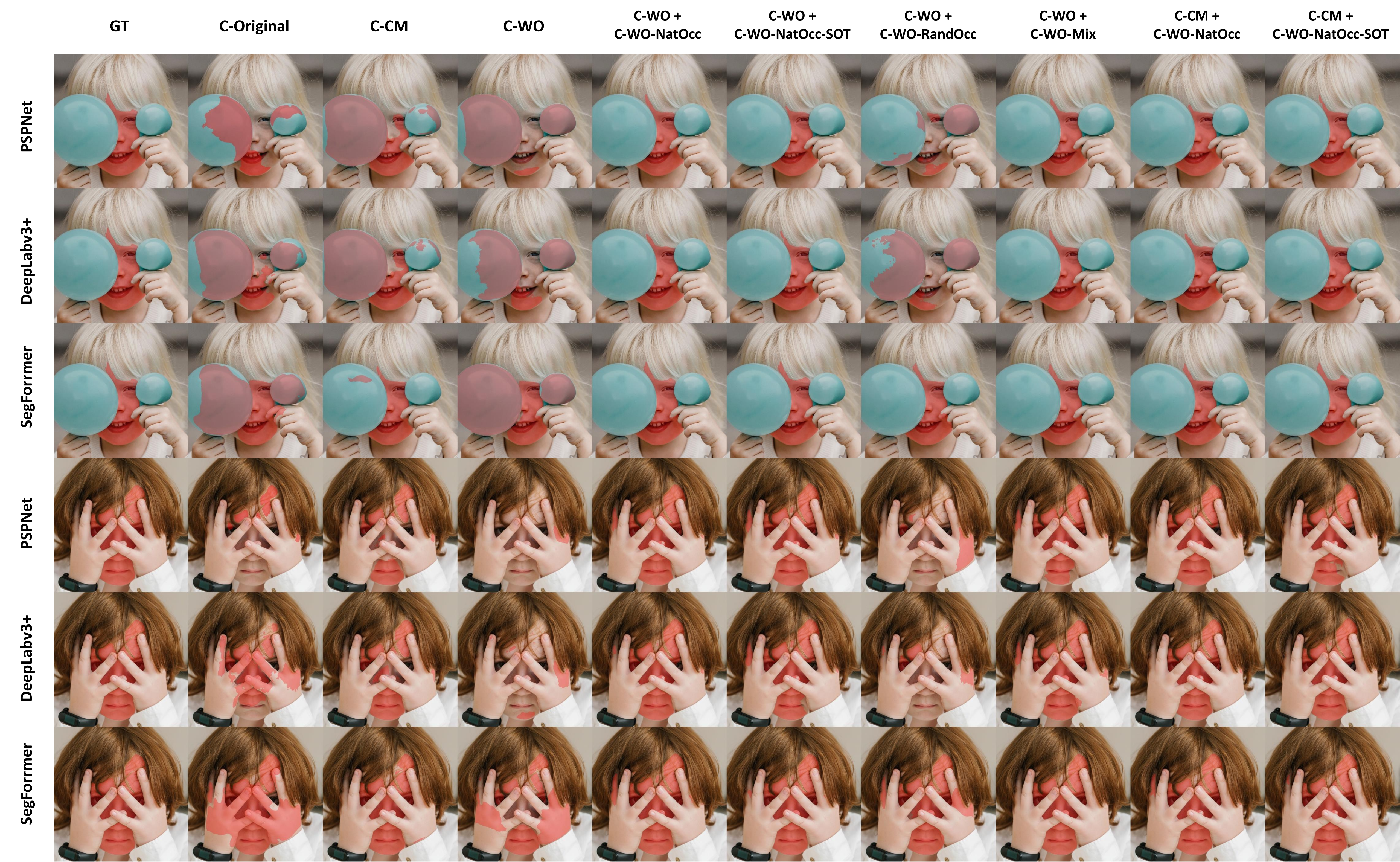}
\vspace{-4mm}

\caption{Occlusion-aware face segmentation results of PSPNet \cite{zhao2017pspnet}, DeepLabv3+ \cite{deeplabv3plus2018} and SegFormer \cite{xie2021segformer} trained on different variations of datasets. Overall, the models trained with our NatOcc datasets obtain comparable or even better results than the models trained with real-occluded dataset (C-CM). SegFormer \cite{xie2021segformer} trained with our RandOcc shows a huge improvement over C-Original and C-WO. } 

\label{fig:qualitative}
\vspace{-3mm}

\end{figure*}

\noindent \textbf{NatOcc.} Table \ref{tab:result} shows that the models trained with our NatOcc dataset (C-WO-NatOcc-SOT, C-WO-NatOcc) performed at a level or better than the real-world occluded face dataset (C-CM), demonstrating the effectiveness of our method, especially with the CNN models (PSPNet \cite{zhao2017pspnet} and DeepLabv3+ \cite{deeplabv3plus2018}). As shown in Figure \ref{fig:qualitative}, the CNN models trained with NatOcc datasets can segment the faces more accurately than the ones trained with the C-CM. Adding C-CM to the NatOcc dataset further enhanced the model's performance. In particular, SegFormer \cite{xie2021segformer} trained on C-WO-NatOcc-SOT and C-CM achieved the highest score of 98.02. Although our synthetic datasets do not cover most of the occlusions (\emph{e.g.}, glasses, face masks, camera, and food), the models trained with our NatOcc dataset can generalize well to these unseen occluded faces. We include more results in \emph{Appendix}.


\noindent \textbf{RandOcc.} The RandOcc dataset (C-WO-RandOcc) brought huge improvement over the one without occlusions (C-WO), and it was not far behind the real-world occluded dataset (C-CM). Figure \ref{fig:qualitative} shows that SegFormer~\cite{xie2021segformer} trained with RandOcc datasets was able to generalize to unseen occlusions despite the unnatural occluders of RandOcc datasets (see Figure \ref{fig:RandOcc}). The advancement in deep learning models has made this type of general data generation practical and possible. C-WO-Mix, the mixture of RandOcc and NatOcc further improved the performance closer to C-CM. Further exploration can be done in future work to create a more robust and effective general synthetic occluded data generation method that can save human labor and time preparing synthetic data for every application. 

\begin{figure}[t]

\centering
\includegraphics[width=1\linewidth]{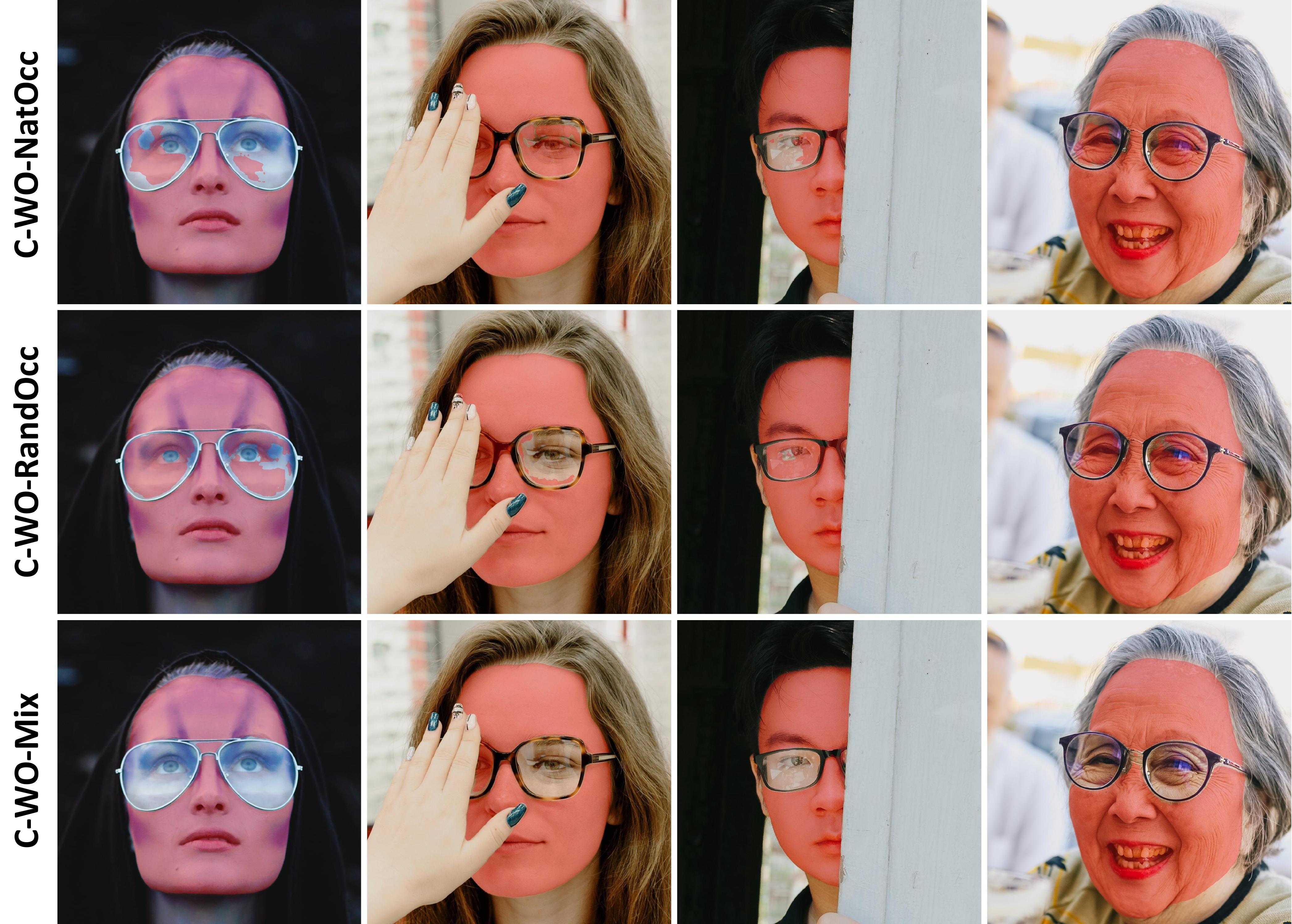}

\caption{The visual results on transparent/translucent occluders of SegFormer\cite{xie2021segformer} trained on C-WO-NatOcc, C-WO-RandOcc and C-WO-Mix. Models trained on C-WO-NatOcc and C-WO-RandOcc failed to detect transparent glasses as occlusion accurately. In contrast, the model on C-WO-Mix can detect transparent glasses as occlusions, proving that RandOcc is complementary to NatOcc to detect transparent/translucent objects.} 
\label{fig:transparent_result}
\vspace{-2mm}

\end{figure}

\noindent \textbf{Hard Occluders (Transparent/Translucent).} SegFormer~\cite{xie2021segformer} trained on the dataset without any real transparent/translucent occluders (C-WO-NatOcc) failed to accurately detect transparent glasses, as shown in Figure~\ref{fig:transparent_result}. Although C-WO-RandOcc contains synthetic transparent/translucent occluders generated by RandOcc in Section~\ref{sec:simulate_transparency}, the model trained with C-WO-RandOcc alone was not able to detect transparent glasses as occlusion. However, the model trained on C-WO-Mix, \emph{i.e.}, a mixture of the datasets generated by NatOcc and RandOcc, was able to detect transparent glasses as occlusion, as shown in Figure \ref{fig:transparent_result}, demonstrating the possibility to simulate transparent/translucent objects. The examples in Figure \ref{fig:transparent_result} show that RandOcc is complementary to NatOcc to potentially detect transparent/translucent objects. 


\noindent \textbf{Impact of an Unclean Dataset.} Table~\ref{tab:result} and Figure~\ref{fig:qualitative} show that the C-Original that has incorrect masks (ignored some occlusions) performed poorly and surprisingly worse than the C-WO, which does not have any occluded faces. This indicates that unclean datasets have a significant negative impact on the model's performance even with SegFormer~\cite{xie2021segformer}, the better deep learning model. Thus, it shows the importance of clean data. 

\noindent \textbf{Class Imbalance.} Occlusions would increase the already large amount of background pixels in an image. Therefore, the IoU of background significantly affects the score of the mIoU, resulting in the high mIoU of the dataset with incorrect annotations (C-Original) and the dataset without occlusions (C-WO). For instance, the background and face IoU of PSPNet \cite{zhao2017pspnet} trained on C-original are 95.04 and 83.99, respectively, and this result in a mIoU of 89.52. Despite having high mIOU, PSPNet trained on C-original perform badly on occlusion-aware face segmentation as shown in Figure~\ref{fig:qualitative}. Therefore, better metrics such as frequency weighted IoU can be explored in future work to better evaluate the model performance.

\noindent \textbf{Impact of Color Transfer.} Despite the minor differences between the model performance trained with C-WO-NatOcc and C-WO-NatOcc-SOT (\emph{i.e.}, with or without color transfer), no conclusion can be made on which one is better. Without a hand-occluded face dataset with annotation masks, the effectiveness of color transfer cannot be evaluated fairly. In reality, faces might be occluded by the hands of another person, which are very different in color from that of the faces. A mixture of hands with and without color transfer might be able to complement each other and achieve higher performance.

\noindent \textbf{Robustness Analysis.} The robustness of the trained models was evaluated on real-world occluded faces in the wild, specifically COFW (Train)~\cite{dataset_cofw} and RealOcc-Wild. Overall, the models trained with our synthetic datasets performed better and can boost the performance of the real-world occluded face dataset (C-CM). In addition, the results on CelebAMask-HQ-WO (Test) shows that despite the improvement on segmenting occluded faces, NatOcc and RandOcc datasets did not bring any negative side effect on segmenting non-occluded faces. Please refer to \emph{Appendix} for additional test results.


\section{Conclusion and Future Work}

In this paper, we proposed NatOcc and RandOcc, two occlusion generation methods that are proven to be effective for occlusion-aware face segmentation, even for unseen occlusions.
Besides, we contributed the corrected masks and new categories of CelebAMask-HQ~\cite{dataset_CelebAMask-HQ}.
To facilitate model evaluation, we introduce two high-resolution real-world occluded face datasets, RealOcc and RealOcc-Wild.
Further, we benchmark several representative baselines and provide insights for future exploration.
As for future work, devising more advanced techniques to produce higher-quality synthetic data to better simulate real-world data could be interesting.
Moreover, improving generalization of synthetic data can be further studied.
In addition, our benchmark can be enriched with higher-quality synthetic data and more baselines for a more comprehensive analysis.
Last but not least, practical applications of occlusion-aware face segmentation in face-related tasks (\emph{e.g.}, face recognition and face-swapping) could be further explored.

\vspace{0.1cm}
\noindent
\textbf{Acknowledgments.} This study is supported under the RIE2020 Industry Alignment Fund – Industry Collaboration Projects (IAF-ICP) Funding Initiative, as well as cash and in-kind contribution from the industry partner(s).





\newpage

{\small
\bibliographystyle{./ieee_fullname}
\bibliography{./egbib}
}
\clearpage

\noindent \textbf{\large Appendix}
\appendix

\section{Additional Quantitative Results}
More examples of the generated datasets and quantitative results are presented in this section. 

\noindent \textbf{Non-occluded Face Dataset.} Table \ref{tab:result1} shows the test results on the CelebAMask-HQ-WO (Test), which is a face dataset without any occlusion. This test aims to verify that our generated dataset will not affect the models' performance in segmenting non-occluded faces. The models trained with the synthetic dataset generated by NatOcc and RandOcc are improved compared to the models trained on C-original and C-WO.

\noindent \textbf{Cropped and Aligned COFW Dataset.} Additional test has been carried out on cropped and aligned COFW dataset~\cite{dataset_cofw}. 400 face images was successfully obtained using the same method as FFHQ~\cite{dataset_karras2019stylebased}. Table \ref{tab:result1} shows that the overall performance is better compared to the COFW dataset~\cite{dataset_cofw} without cropped and aligned.

\section{Additional Qualitative Results}
\begin{figure*}[t]
\vspace{-2mm}
\centering
\includegraphics[width=1\linewidth]{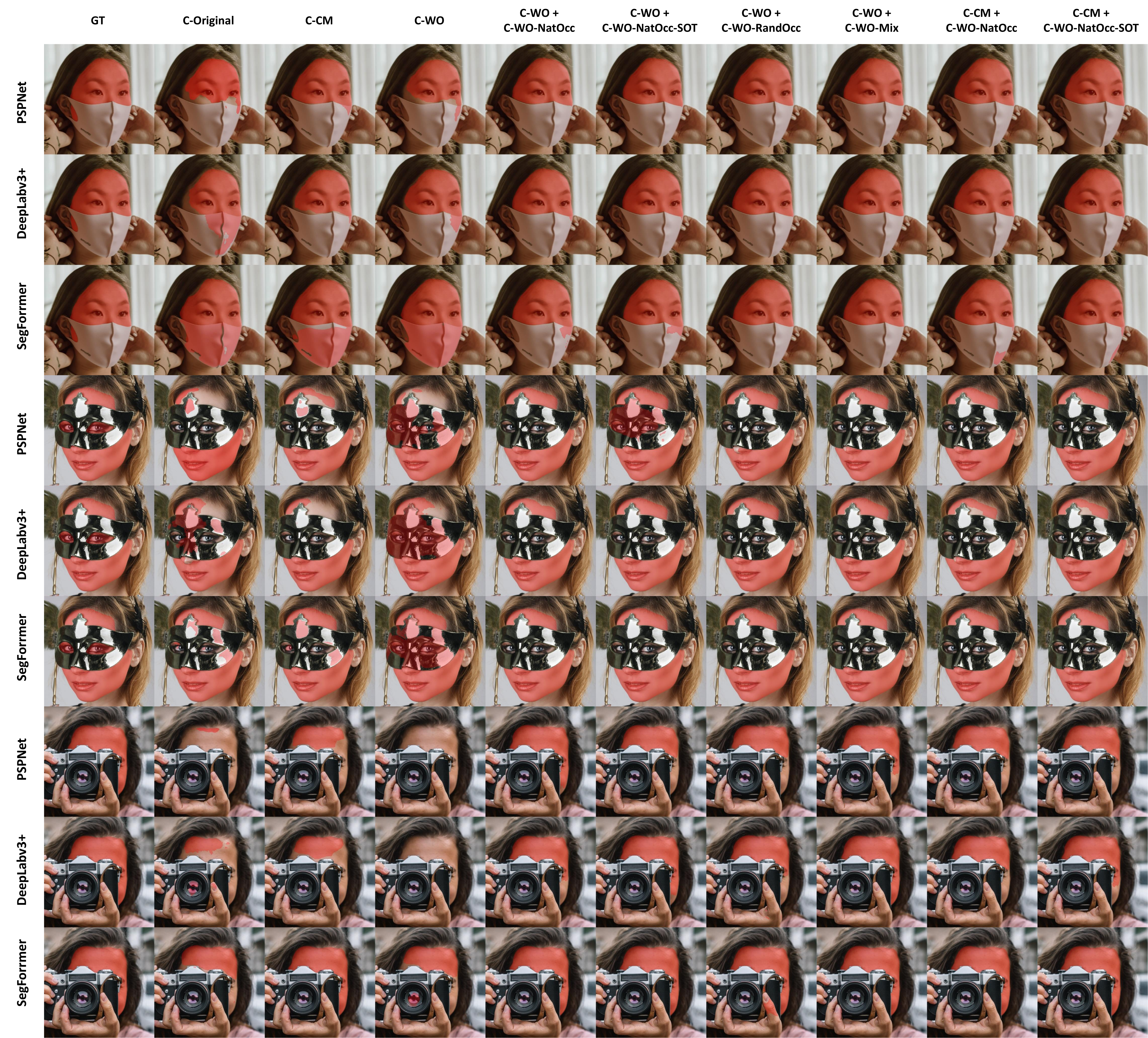}
\vspace{-1mm}

\caption{ Examples of the inference results on hand-occluded faces. NatOcc and RandOcc are effective in simulating real-world occluded datasets. } 

\label{fig:inference1}
\vspace{-5mm}

\end{figure*}

\begin{figure*}[t]
\vspace{-2mm}
\centering
\includegraphics[width=1\linewidth]{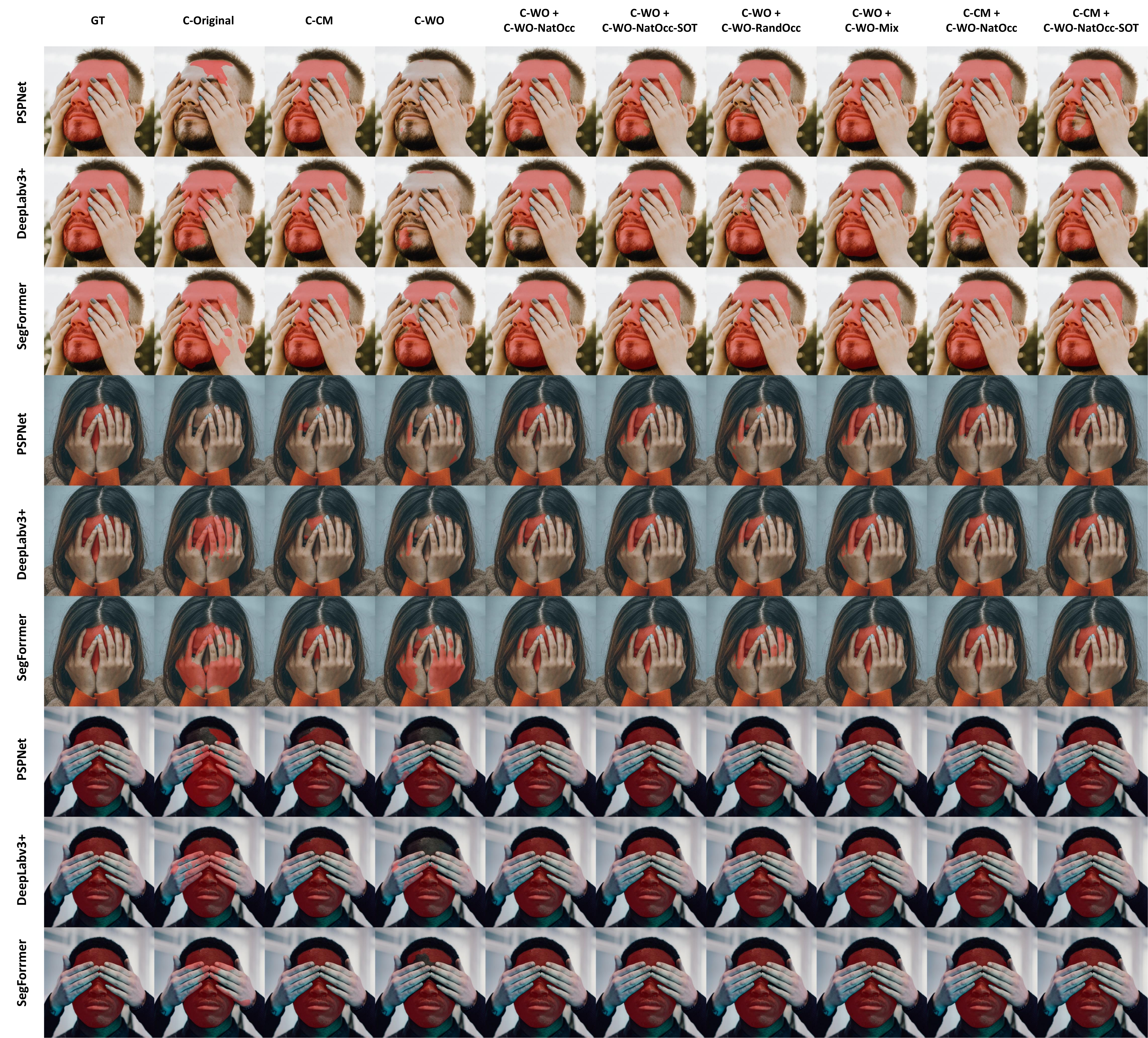}
\vspace{-1mm}

\caption{ Examples of the inference results on objects-occluded faces. NatOcc and RandOcc are effective in simulating real-world occluded datasets.  } 

\label{fig:inference2}
\vspace{-5mm}

\end{figure*}

The additional qualitative results in Figure \ref{fig:inference1}
and Figure~\ref{fig:inference2} show that models trained with both NatOcc and RandOcc datasets perform as well as or better than models trained on the real-world dataset (C-WO). In some examples, they are even better than real-world datasets, showing the effectiveness of our data generation methods.

\section{More Examples of NatOcc and RandOcc Datasets}
\begin{figure*}[t]
\vspace{-2mm}
\centering
\includegraphics[width=1\linewidth]{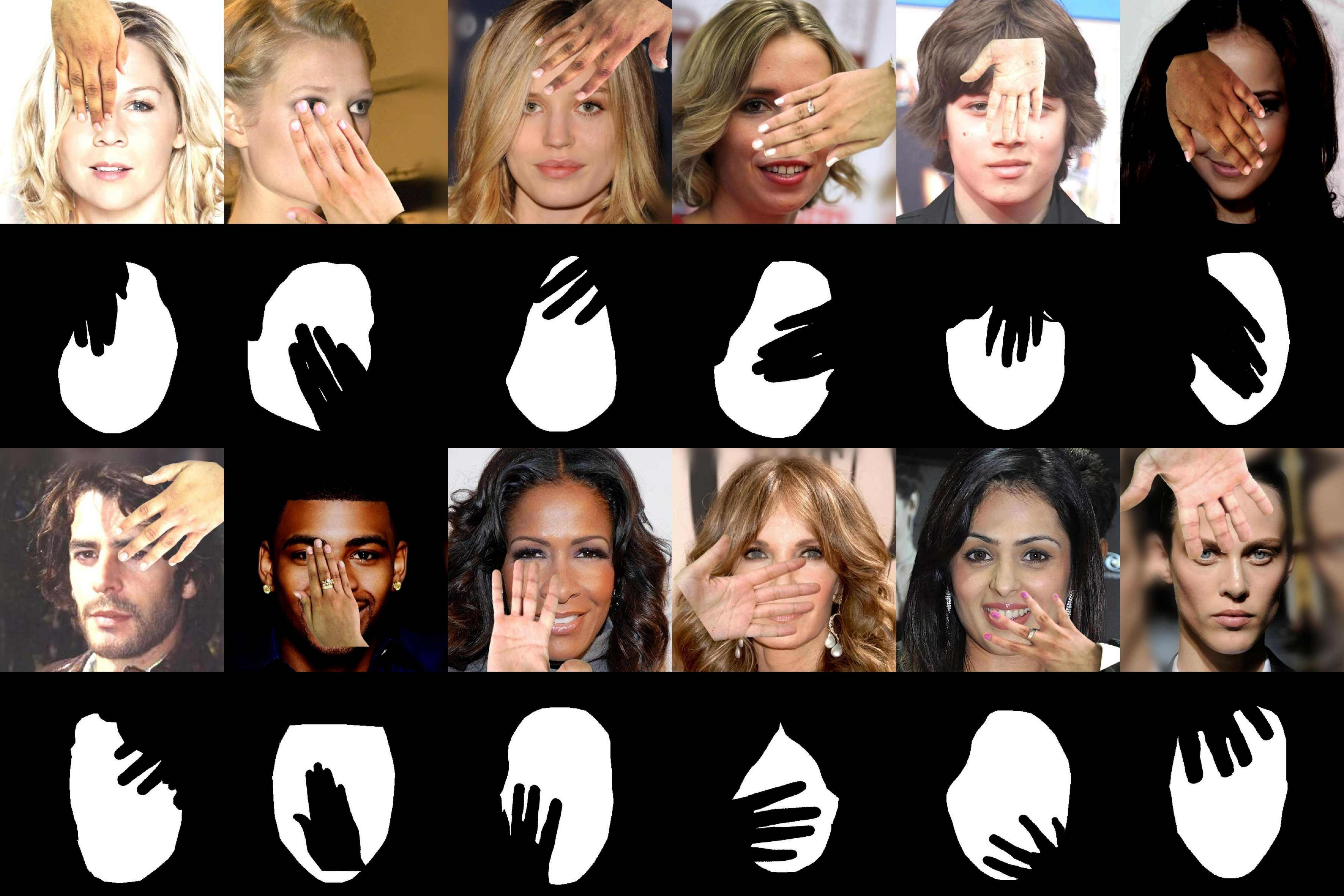}
\vspace{-1mm}

\caption{ Examples of the hand-occluded faces generated by NatOcc without color transfer. } 

\label{fig:natocc1}
\vspace{-5mm}

\end{figure*}

\begin{figure*}[t]
\vspace{-2mm}
\centering
\includegraphics[width=1\linewidth]{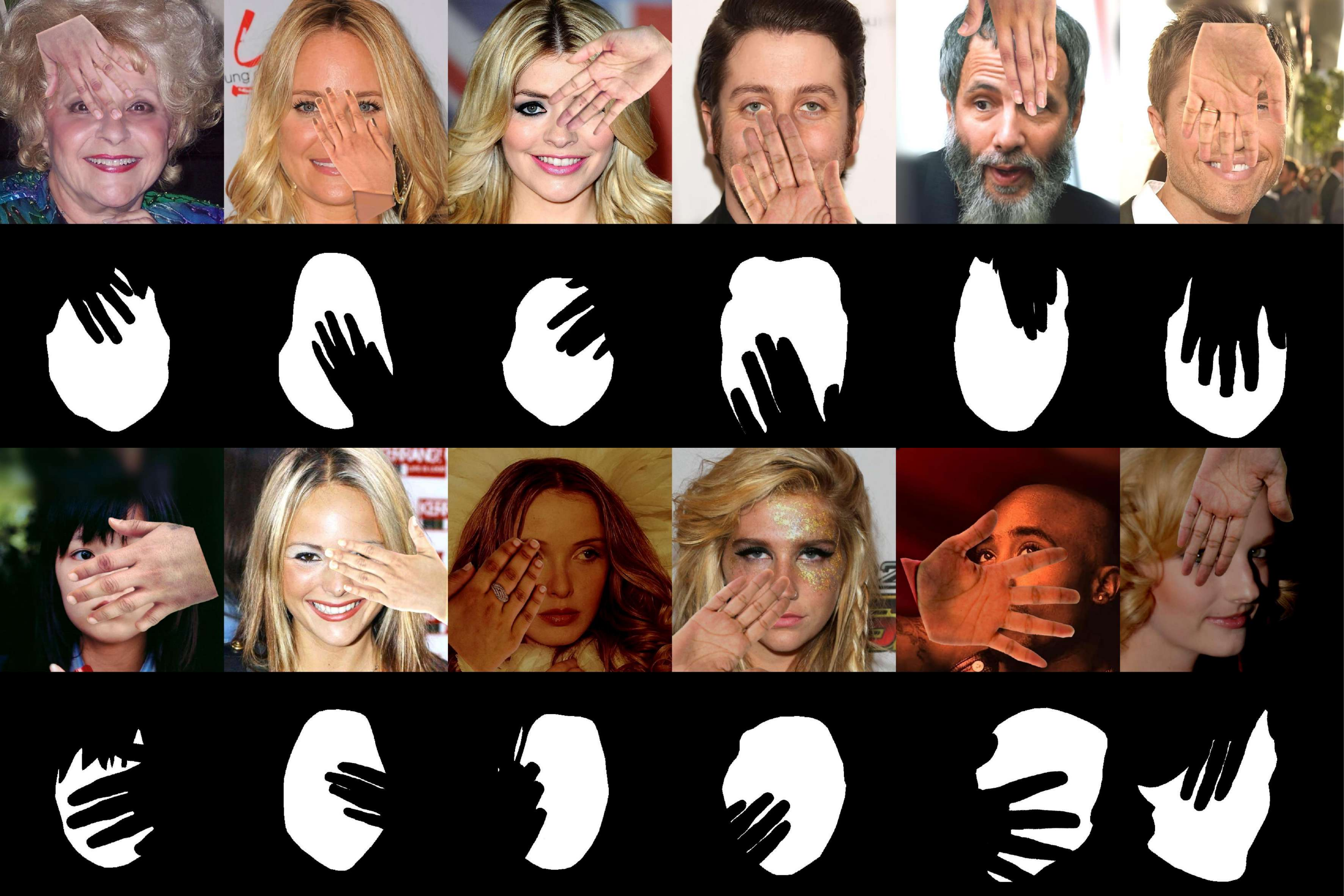}
\vspace{-1mm}

\caption{Examples of the hand-occluded faces generated by NatOcc with color transfer.}

\label{fig:natocc2}
\vspace{-5mm}

\end{figure*}

\begin{figure*}[t]
\vspace{-2mm}
\centering
\includegraphics[width=1\linewidth]{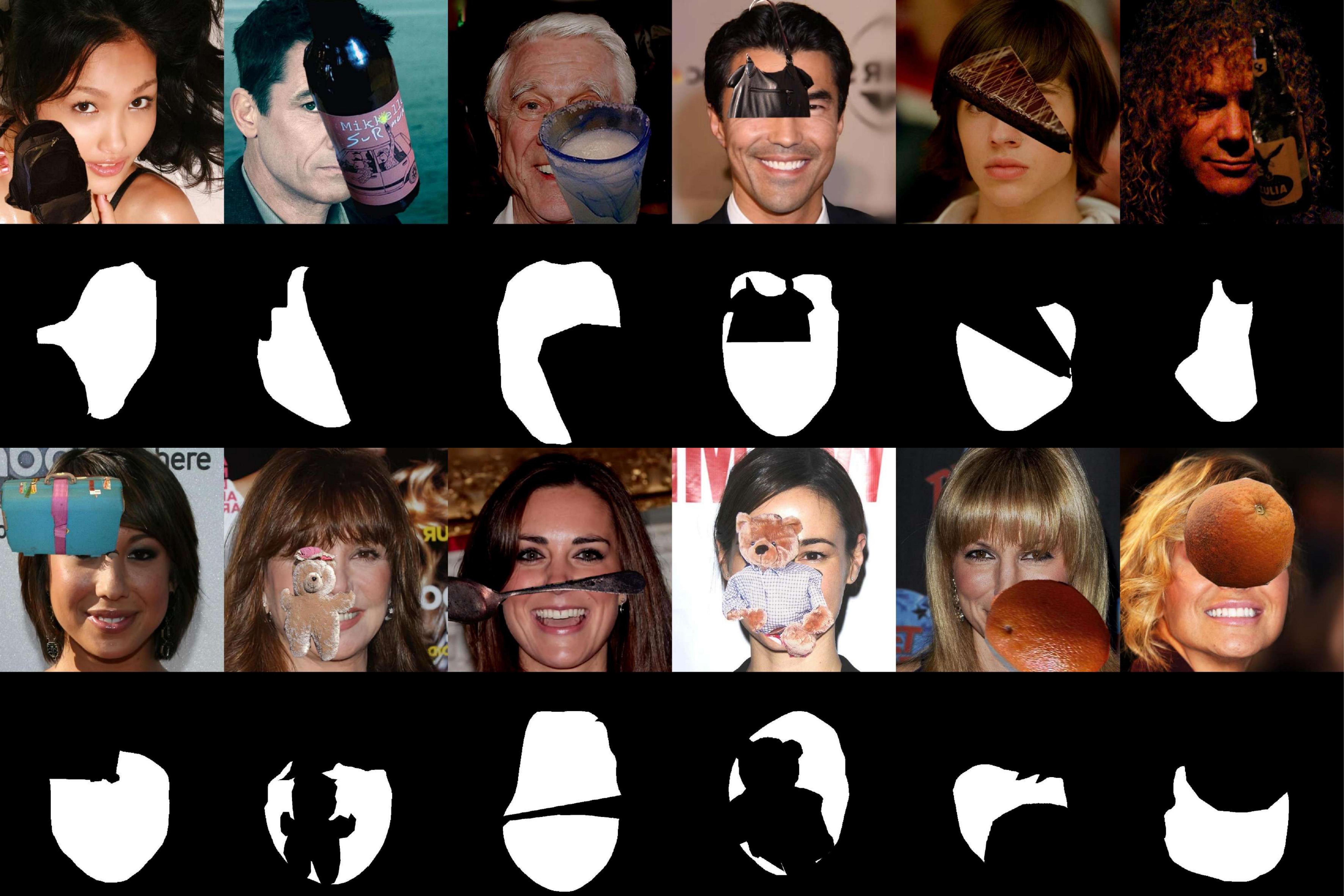}
\vspace{-1mm}

\caption{Examples of the COCO \cite{dataset_lin2015microsoft} objects occluded faces generated by NatOcc without color transfer. } 

\label{fig:natocc3}
\vspace{-5mm}

\end{figure*}

\begin{figure*}[t]
\vspace{-2mm}
\centering
\includegraphics[width=1\linewidth]{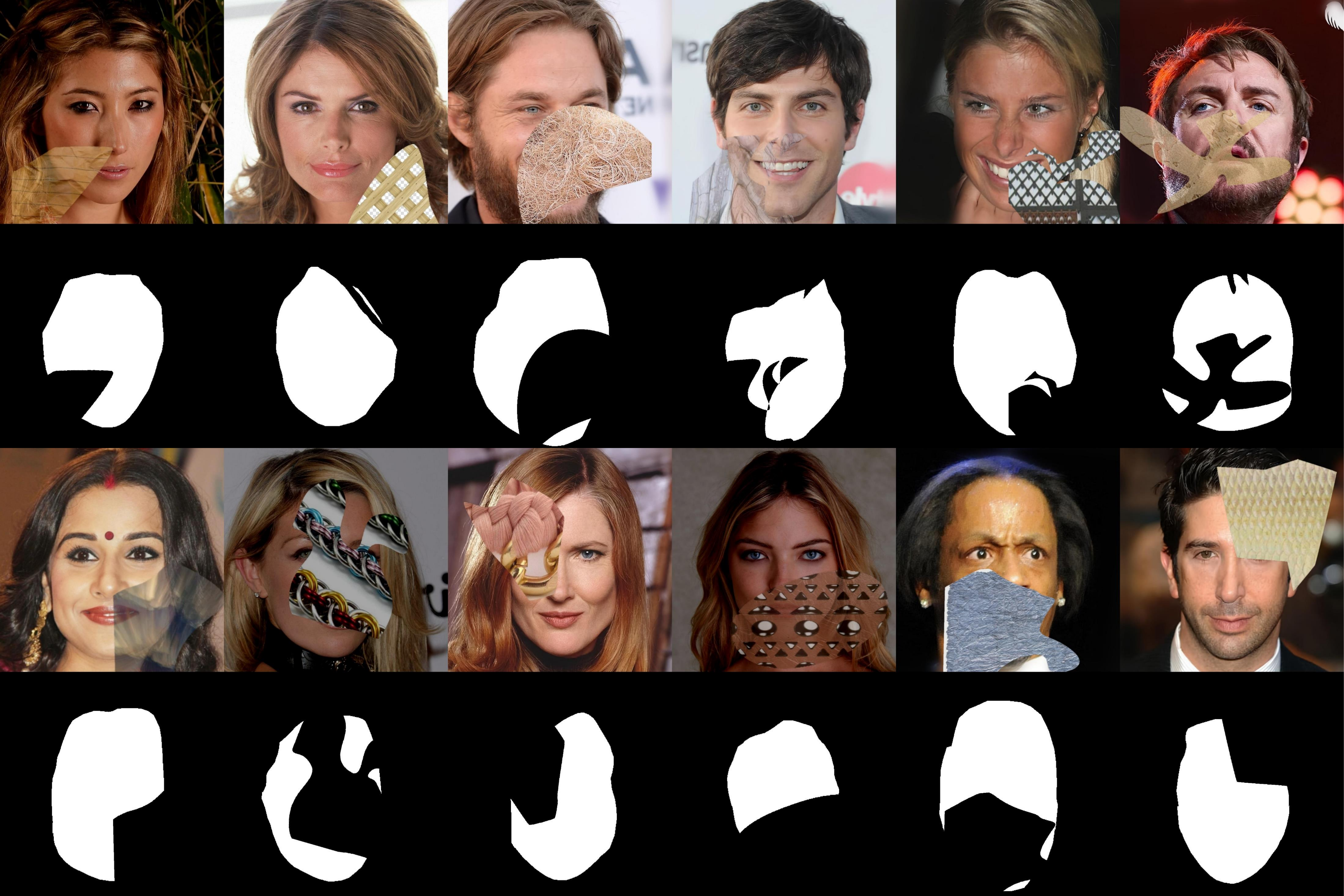}
\vspace{-1mm}

\caption{Examples of the images generated by RandOcc by overlaying random shape with random transparency and texture from DTD \cite{dataset_dtd}. } 

\label{fig:rand1}
\vspace{-5mm}

\end{figure*}

Additional examples of the NatOcc dataset are shown in Figure \ref{fig:natocc1}, Figure \ref{fig:natocc2}, and Figure \ref{fig:natocc3}. Figure \ref{fig:natocc1} shows the hand-occluded faces without color transfer while Figure~\ref{fig:natocc2} shows the hand-occluded faces with color transfer. Figure \ref{fig:natocc3} shows the COCO \cite{dataset_lin2015microsoft} objects-occluded faces after image harmonization. Moreover, some examples of the RandOcc dataset are shown in Figure \ref{fig:rand1}.

\section{More Examples of Validation Sets}
\begin{figure*}[t]
\vspace{-2mm}
\centering
\includegraphics[width=1\linewidth]{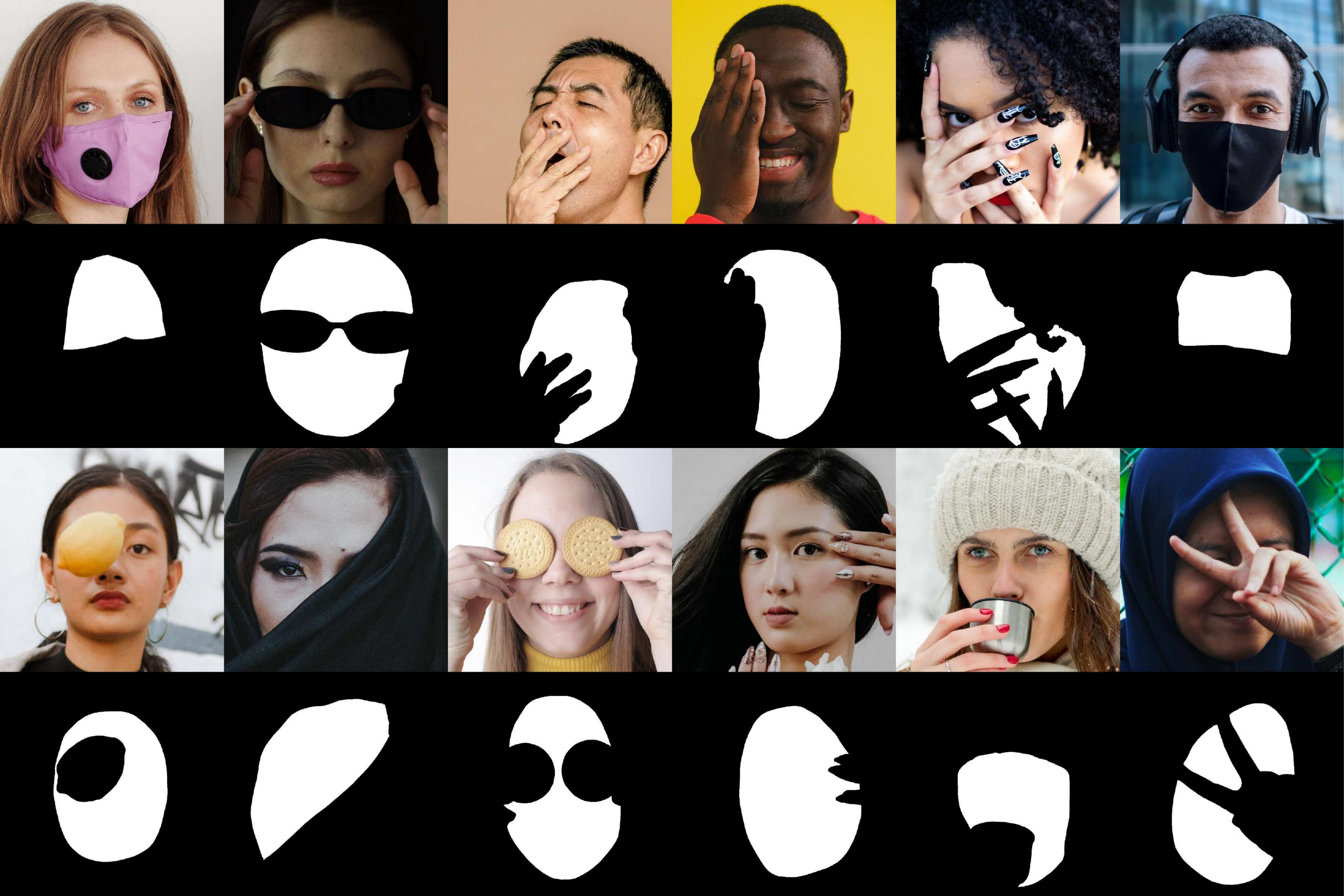}
\vspace{-1mm}

\caption{Examples of the RealOcc, aligned and cropped real-world occluded faces. } 

\label{fig:realocc}
\vspace{-5mm}

\end{figure*}

\begin{figure*}[t]
\vspace{-2mm}
\centering
\includegraphics[width=1\linewidth]{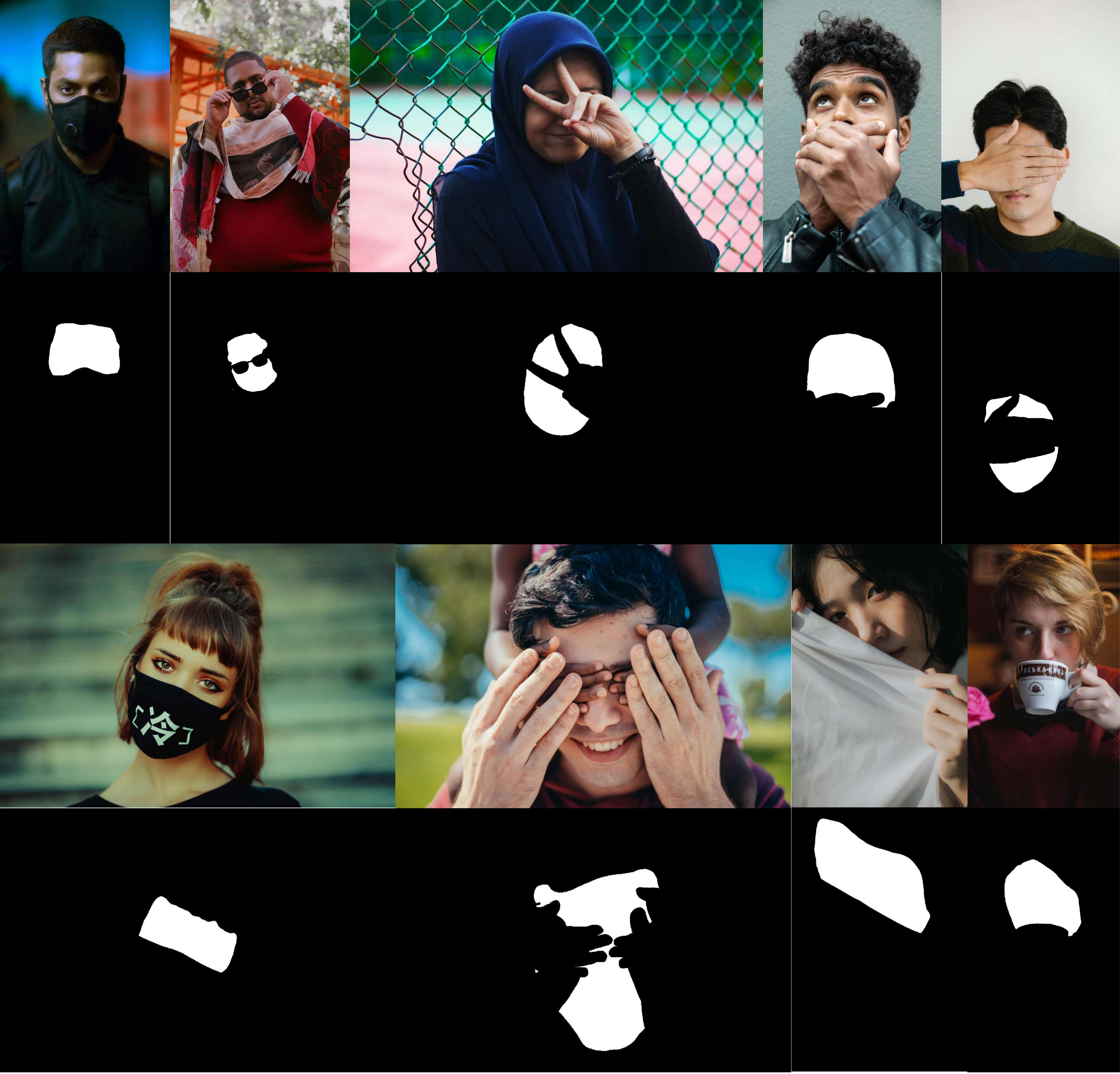}
\vspace{-1mm}

\caption{ Examples of the RealOcc-Wild, real-world occluded faces in the wild.} 

\label{fig:realoccwild}
\vspace{-5mm}

\end{figure*}

Figure \ref{fig:realocc} shows additional examples of the RealOcc while Figure \ref{fig:realoccwild} shows some examples of occluded faces in the wild.

\begin{table*}[ht]
    \centering
    \caption{ \textbf{Additional quantitative test results:} PSPNet \cite{zhao2017pspnet}, DeepLabv3+ \cite{deeplabv3plus2018}, and SegFormer \cite{xie2021segformer} with different combination of datasets. The best results for each validation set are marked in bold. The metrics are mIoU (higher is better).
        }
    \resizebox{1\textwidth}{!}
    {
    \begin{tabular}{lcccccccccc}
    \toprule
    & Quantity &&\multicolumn{3}{c}{\textbf{CelebAMask-HQ-WO (Test)} (mIoU)} &
    &\multicolumn{3}{c}{\textbf{COFW (Train) (cropped and aligend)} (mIoU)} \\
    \cmidrule{4-6} \cmidrule{8-10} 
    & & &  PSPNet & DeepLabv3+  & SegFormer  
       & & PSPNet & DeepLabv3+  & SegFormer   
     \\
    \midrule
     C-Original & 29,200                       &&97.71 &97.23&97.18     && 93.21&92.37&92.87 \\
     C-CM & 29,200                    &&  \textbf{97.78} &\textbf{97.79}& \textbf{97.88}            && 95.34&\textbf{95.32}& \textbf{95.62} \\
     C-WO  & 24,602                           && 97.66 & 97.70&97.84    && 92.88&92.74& 93.54\\               
     C-WO + C-WO-NatOcc & 24,602 + 49,204        && 97.77& 97.76&97.86  && 94.45&94.46&94.87\\ 
     C-WO + C-WO-NatOcc-SOT & 24,602 + 49,204    &&  97.71& 97.77&97.87     &&94.61&94.47&94.63\\

     C-WO + C-WO-RandOcc & 24,602 + 49,204       && 97.68&97.76 &97.83     &&   93.97&93.83 & 94.19\\
     C-WO + C-WO-Mix & 24,602 + 49,204          &&  97.70&97.76 &97.76  &&         93.92&94.55  & 94.67\\
      C-CM + C-WO-NatOcc & 29,200 + 49,204  && 97.74&\textbf{97.79}&97.87        && \textbf{95.38}&\textbf{95.32}&95.53\\ 
      C-CM + C-WO-NatOcc-SOT & 29,200 + 49,204  && \textbf{97.78}&97.76&97.85  && 95.26&95.23&95.46\\ 
    \bottomrule
    \end{tabular}
    }

  \label{tab:result1}

\end{table*}

\end{document}